\documentclass[runningheads]{llncs}
\usepackage{graphicx}

\usepackage{tikz}
\usepackage{comment}
\usepackage{amsmath,amssymb} 
\usepackage{color}
\usepackage[accsupp]{axessibility}  

\usepackage{nicefrac}
\usepackage{wrapfig}

\usepackage{bm}
\usepackage{bbm}
\usepackage{mathrsfs}
\usepackage{multirow}
\usepackage{url}
\usepackage{caption}
\usepackage{array}
\usepackage{pifont}
\usepackage{color,xcolor}
\usepackage{algorithm, algpseudocode}
\definecolor{mydarkblue}{rgb}{0,0.08,0.45}

\DeclareMathOperator*{\argmax}{arg\,max} 
\DeclareMathOperator*{\argmin}{arg\,min} 
\newcommand{\tabincell}[2]{\begin{tabular}{@{}#1@{}}#2\end{tabular}} 

\begin{document}
\pagestyle{headings}
\mainmatter
\def\ECCVSubNumber{4374}  

\title{
Boosting Transferability of Targeted Adversarial Examples via Hierarchical Generative Networks} 

\titlerunning{ECCV-22 submission ID \ECCVSubNumber} 
\authorrunning{ECCV-22 submission ID \ECCVSubNumber} 
\author{Anonymous ECCV submission}
\institute{Paper ID \ECCVSubNumber}

\titlerunning{Targeted Adversarial
Examples via Hierarchical Generative Networks}
%
\author{Xiao Yang$^{1}$ ~~ Yinpeng Dong$^{1,2}$ ~~ Tianyu Pang$^{3}$ ~~
Hang Su$^{1,4}$ ~~
Jun Zhu$^{1,2,4}$\thanks{corresponding author.}}
\authorrunning{X.Yang et al.}
%
\institute{Dept. of Comp. Sci. and Tech., Institute for AI, Tsinghua-Bosch Joint ML Center, THBI Lab, BNRist Center, Tsinghua University$^{1}$ \\ RealAI$^{2}$ \qquad
Sea AI Lab, Singapore$^{3}$ \\
{Peng Cheng Laboratory; Pazhou Laboratory (Huangpu), Guangzhou, China}$^{4}$ \\
\email{\small{\{yangxiao19, dyp17\}@mails.tsinghua.edu.cn}, \small{ tianyupang@sea.com}, \small{\{suhangss, dcszj\}@tsinghua.edu.cn}
}
}
\maketitle

\begin{abstract}
Transfer-based adversarial attacks can evaluate model robustness in the black-box setting. Several methods have demonstrated impressive untargeted transferability, however, it is still challenging to efficiently produce \emph{targeted} transferability. To this end, we develop a simple yet effective framework to craft targeted transfer-based adversarial examples, applying a hierarchical generative network. In particular, we contribute to amortized designs that well adapt to multi-class targeted attacks. Extensive experiments on ImageNet show that our method improves the success rates of targeted black-box attacks by a significant margin over the existing methods --- it reaches an average success rate of 29.1\% against six diverse models based only on one substitute white-box model, which significantly outperforms the state-of-the-art gradient-based attack methods. Moreover, the proposed method is also more efficient beyond an order of magnitude than gradient-based methods.
\end{abstract}

\section{Introduction}

Recent progress in adversarial machine learning demonstrates that deep neural networks (DNNs) are highly vulnerable to adversarial examples~\cite{goodfellow2014explaining,szegedy2013intriguing}, which are maliciously generated to mislead a model to produce incorrect predictions. It has been demonstrated that adversarial examples possess an intriguing property of transferability~\cite{wu2020skip,huang2019enhancing,demontis2019adversarial} --- the adversarial examples crafted for a white-box model can also mislead other unknown models, making \emph{black-box attacks} feasible. The threats of adversarial examples have raised severe concerns in numerous security-sensitive applications, such as autonomous driving~\cite{eykholt2018robust} and face recognition~\cite{yang2020towards,yang2020design,yang2022controllable}.

Tremendous efforts have been made to develop more effective black-box attacking methods based on transferability, since they can serve as an important surrogate to evaluate the model robustness in real-world scenarios~\cite{liu2016delving,Dong2017}. The current methods have achieved impressive
performance of untargeted black-box attacks, intending to cause misclassification of the black-box models. However, \emph{targeted} black-box attacks, aiming at misleading the black-box models by outputting the adversary-desired target class, perform unsatisfactorily or require computation scaling with number of classes~\cite{dong2019benchmarking,zhang2020understanding}. Technically, the inefficiency of targeted adversarial attacks could result in an over-estimation of model robustness under the challenging black-box attack setting~\cite{hendrycks2021unsolved}.

Existing efforts on targeted black-box attacks can be categorized as \emph{instance-specific} and \emph{instance-agnostic} attacks. Specifically, the instance-specific attack methods~\cite{Goodfellow-et-al2016,Moosavidezfooli2016,Kurakin2016,Dong2017,li2020towards} craft adversarial examples by performing gradient updates iteratively, which obtain unsatisfactory performance for targeted black-box attacks due to easy overfitting to a white-box model~\cite{Dong2017,xie2019improving}. Recently, \cite{zhao2021success} propose several improvements for instance-specific targeted attacks, thus we treat the method in \cite{zhao2021success} as one of the strong instance-specific baselines compared in our experiments.

On the other hand, the instance-agnostic attack methods learn a universal perturbation~\cite{zhang2020understanding} or a universal function~\cite{song2018constructing,naseer2019cross} on the data distribution independent of specific instances. They can promote more general and transferable adversarial examples since the universal perturbation or function can alleviate the data-specific overfitting problem by training on an unlabeled dataset. 
CD-AP~\cite{naseer2019cross}, as one of the effective instance-agnostic methods, adopts a generative model as a universal function to obtain an acceptable performance when facing one specified target class. However, CD-AP needs to learn a generative model for each target class while performing multi-target attack~\cite{han2019once}, i.e., crafting adversarial examples targeted at different classes. Thus it is not scalable to the increasing number of targets such as hundreds of classes, limiting practical efficiency.

To address the aforementioned issues and develop a targeted black-box attack in the practical scenario, in this paper we propose a conditional generative model as the universal adversarial function to craft adversarial perturbations. 
Thus we can craft adversarial perturbations targeted at different classes, using a single model backbone with different class embeddings. The proposed generative method is simple yet practical to obtain superior performance of targeted black-box attacks, meanwhile with two technical improvements including (\romannumeral1) {smooth projection mechanism} that better helps the generator probe targeted semantic knowledge from the classifier; (\romannumeral2) {adaptive Gaussian smoothing} with the focus of making generated results obtain adaptive ability against adversarially trained models. Therefore, our approach have several advantages over existing generative attacks~\cite{naseer2019cross,poursaeed2018generative,reddy2018ask}, as described in the followings.

\textbf{One model for multiple target classes.} The previous generative methods~\cite{naseer2019cross,poursaeed2018generative} require costly training $N$ models while performing a multi-target attack with $N$ classes. However, ours only trains one model and reaches an average success rate of 51.1\% against six naturally trained models and 36.4\% against three adversarially trained models based only on one substitute white-box model in ImageNet dataset, which outperforms CD-AP by a large margin of 6.0\% and 31.3\%, respectively.

\textbf{Hierarchical partition of classes.} While handling plenty of classes (e.g., 1,000 classes in ImageNet), the effectiveness of generating targeted adversarial examples
will be affected by a single
generative model due to the difficulty of loss convergence in adversarial learning~\cite{xu2019understanding,berthelot2017began}. Thus we train a feasible number of models (e.g., 10$\sim$20 models on ImageNet) to further promote the effectiveness beyond the single model backbone. 
Specifically, each model is learned from a subset of classes specified by a designed hierarchical partition mechanism by considering the diversity property among subsets, for seeking a balance between effectiveness and scalability. 
It reaches an average success rate of 29.1\% against six different models, outperforming the state-of-the-art gradient-based methods by a large margin, based only on one substitute white-box model. Moreover, the proposed method achieves substantial speedup over the mainstream gradient-based methods.

\begin{figure*}[t]
\begin{center}
\includegraphics[width=0.99\linewidth]{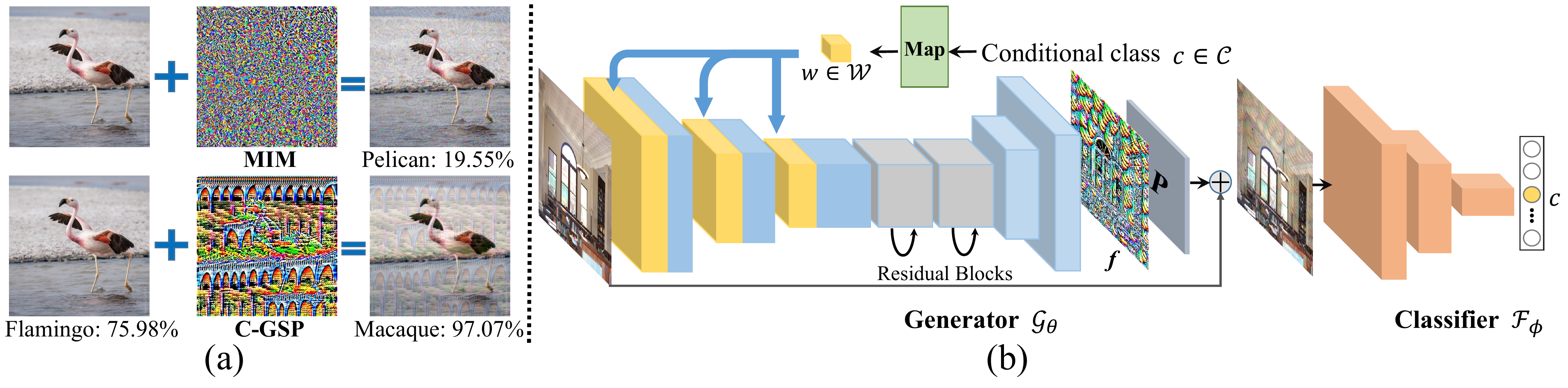}
\end{center}
\caption{(a) shows the targeted adversarial examples crafted by MIM~\cite{Dong2017} and C-GSP given the target class \emph{Viaduct} with the maximum perturbation $\epsilon = 16$. The predicted labels and probabilities are shown by another black-box model. (b) presents an overview of our proposed generative method for crafting C-GSP, including modules of conditional generator and classifier. The generator integrates the image and conditional class vector from Map network into a hidden incorporation. The generator is only trained in the whole pipeline to probe the target boundaries of the classifier.}
\label{fig:model}
\end{figure*}

\textbf{Strong semantic patterns.} We experimentally find that these adversarial perturbations generated by the proposed Conditional Generative models can arise as a result of strong Semantic Pattern (C-GSP) as shown in Fig.~\ref{fig:model}(a). Furthermore, we present more valuable analyses in Sec.~\ref{lab:discuss}, illustrating that the generated semantic pattern itself achieves well-generalizing performance among the different models and is robust to the influence of data. These analyses are very instructive for the understanding and design of adversarial examples.

Technically, our main contributions can be summarized as: 
\begin{itemize}
    \item We propose a simple yet practical conditional generative targeted attack with a scalable hierarchical partition mechanism, 
    which can generate targeted adversarial examples without tuning the parameters. 
    
    \item Extensive experiments demonstrate that our method significantly improves the success rates of targeted black-box attacks over the existing methods.
    \item As a by-product, our baseline experiments provide a systematical evaluation on previous targeted black-box attacks, either instance-specific or instance-agnostic, on the ImageNet dataset with plenty of classes and face recognition.
\end{itemize}

\section{Related Work}

In this section, we review related work on adversarial attacks belonging to different types.

\textbf{Instance-specific attacks.} Some recent works~\cite{Moosavidezfooli2016,zhao2021success} adopt gradient-based optimization methods to generate the data-dependent perturbations. MIM~\cite{Dong2017} introduces the momentum term into the iterative attack process to improve the black-box transferability. DIM~\cite{xie2019improving} and TI~\cite{dong2019evading} aim to achieve the better transferability by input or gradient diversity. Recent works~\cite{inkawhich2020perturbing,inkawhich2020transferable} also attempt to costly train the multiple auxiliary classifiers to improve the black-box performance of iterative methods. In contrast, we improve the transferability performance over instance-specific methods simultaneously with the inference-time efficiency.

\textbf{Instance-agnostic attacks.} Different from instance-specific attacks, instance-agnostic attacks belong to image-independent (universal) methods. The first pipeline is to learn a universal perturbation. UAP~\cite{moosavi2017universal} proposes to fool a model by adding a learned universal noise vector.
Another pipeline of attacks introduces learned generative models to craft adversarial examples. 
GAP~\cite{poursaeed2018generative} and AAA~\cite{reddy2018ask} craft adversarial perturbations based on target data directly and compress impressions, respectively. Previous methods, including universal perturbation and function, require costly training the same number of models for multiple target classes. Our method is capable of simultaneously generating adversarial samples for specifying multiple targets with better attack performance.

\textbf{Multi-target attacks.} Instance-specific attacks have the ability for specifying any target in the optimization phase. As elaborated in the introduction, these methods have degraded transferability and time-consuming iterative procedures. MAN~\cite{han2019once} trains a generative model in the ImageNet under the constraint of $\ell_{2}$ norm to explore the targeted attacks, which specifies all 1,000 categories from ImageNet for seeking extreme speed and storage. However, MAN does not fully compare multi-target black-box performance with previous instance-specific or instance-agnostic attacks, and the authors also claim that too many categories make it hard to transfer to another model. Recent approaches~\cite{zhang2020understanding,naseer2021generating} reveal better single-target transferability by learning universal perturbation or function, whereas they require to train multiple times while specifying multiple targets. As a comparison, 
our method can generate adversarial samples for specifying multiple targets, meanwhile generated strong semantic patterns can outperform existing attacks by a significant margin.

\section{Method}
\label{sec:method}

In this section, we introduce a conditional generative model to learn a universal adversarial function, which can achieve effective multi-target black-box attacks. While handing plenty of classes, we design a hierarchical partition mechanism to make the generative model capable of specifying any target class under a feasible number of models, regarding both the effectiveness and scalability.

\subsection{Problem Formulation}

We use $\bm{x}_{s}$ to denote an input image belonging to an unlabeled training set $\mathcal{X}_{s} \subset \mathbb{R}^{d}$, and use $c \in \mathcal{C}$ to denote a specific target class. Let $\mathcal{F}_{\phi}: \mathcal{X}_{s} \rightarrow \mathbb{R}^{K}$ denote a classification network that outputs a class probability vector with $K$ classes. To craft a targeted adversarial example $\bm{x}_{s}^*$ from a real example $\bm{x}_{s}$, the targeted attack aims to fool the classifier $\mathcal{F}_{\phi}$ by outputting a specific label $c$ as $\argmax_{i\in  \mathcal{C}}{\mathcal{F}_{\phi}(\bm{x}_{s}^{*})}_{i} = c$, meanwhile the $\ell_{\infty}$ norm of the adversarial perturbation is required to be no more than a threshold $\epsilon$ as $\|\bm{x}_{s}^{*}-\bm{x}_{s}\|_{\infty}\leq\epsilon$.

Although some generative methods~\cite{poursaeed2018generative,naseer2019cross} can learn targeted adversarial perturbation, they do not take into account the effectiveness of multi-target generation, thus leading to inconvenience. 
To make the generative model learn how to specify multiple targets, we propose a conditional generative network $\mathcal{G}_{\theta}$ that effectively crafts multi-target adversarial perturbations by modeling class-conditional distribution. 
Different from previous single-target methods~\cite{naseer2019cross,poursaeed2018generative}, the target label $c$ is regarded as a discrete variable rather than a constant. As illustrated in Fig.~\ref{fig:model}(b), our model contains a conditional generator $\mathcal{G}_{\theta}$ and a classification network $\mathcal{F}_{\phi}$ parameterized by $\theta$ and $\phi$, respectively. The conditional generative model $\mathcal{G}_{\theta}: (\mathcal{X}_{s}, \mathcal{C}) \rightarrow \mathcal{P}$ learns a perturbation $\bm{\delta} = \mathcal{G}_{\theta}(\bm{x}_{s}, c) \in \mathcal{P} \subset \mathbb{R}^{d}$ on the training data. The output $\bm{\delta}$ of $\mathcal{G}_{\theta}$ is projected within the fixed $\ell_{\infty}$ norm, thus generating the perturbed image $\bm{x}_{s}^{*} = \bm{x}_{s} + \bm{\delta}$.

Given a pretrained network $\mathcal{F}_{\phi}$ parameterized by  $\phi$, we propose to generate the targeted adversarial perturbations by solving 
\begin{equation}
\begin{split}
    \min\limits_{\theta}&\mathbb{E}_{(\bm{x}_{s}\sim \mathcal{X}_{s}, c\sim \mathcal{C})}[{\mathbb{CE}\big(\mathcal{F}_{\phi}(\mathcal{G}_{\theta}(\bm{x}_{s}, c) + \bm{x}_{s}\big), c)}], \;
  \text{s.t. } \|\mathcal{G}_{\theta}(\bm{x}_{s}, c) \|_{\infty} \leq \epsilon,
\end{split}
\label{eq:target}
\end{equation}
where $\mathbb{CE}$ is the cross-entropy loss. By solving problem~\eqref{eq:target}, we can obtain a targeted conditional generator by minimizing the loss of specific target class in the unlabeled training dataset.
Note that we only optimize the parameter $\theta$ of the generator $\mathcal{G}_{\theta}$ using the training data $\mathcal{X}_{s}$, then the targeted adversarial example $\bm{x}_{t}^{*}$ can be crafted by $\bm{x}_{t}^{*} = \bm{x}_{t} + \mathcal{G}_{\theta}(\bm{x}_{t}, c) $ for any given image $\bm{x}_{t}$ in the test data $\mathcal{X}_{t}$, which only requires an inference for this targeted image $\bm{x}_{t}$. 

We experimentally find that the objective~\eqref{eq:target} can enforce the transferability for the generated perturbation $\bm{\delta}$. A reasonable explanation is that $\bm{\delta}$ can arise as a result of \textbf{strong} and \textbf{well-generalizing} \emph{semantic pattern} inherent to the target class, which is robust to the influence of any training data. In Sec.~\ref{sec:comparison}, we illustrate and corroborate our claim by directly feeding scaled adversarial perturbations\footnote{The perturbation is linearly scaled from [-$\epsilon$, $\epsilon$] to [0, 255].} from different methods into the classifier. Indeed, we find that our semantic pattern can be classified as the target class with a high degree of confidence while the perturbation from MIM~\cite{Dong2017} performs like the noise, meanwhile the scaled semantic pattern performs well transferability in different black-box models. 

\subsection{Network Architecture}
We now present the details of the conditional generative model for targeted attack, as illustrated in Fig.~\ref{fig:model}(b). Specifically, we design a mapping network to generate a target-specific vector in the implicit space of each target and train conditional generator $\mathcal{G}_{\theta}$ to reflect this vector by constantly misleading the classifier $\mathcal{F}_{\phi}$.

\textbf{Mapping network.} Given the one-hot class encoding $\mathbbm{1}_{c} \in \mathbb{R}^{K}$ from target class $c$, the mapping network aims to generate the targeted latent vector $\bm{w} = \mathcal{W}(\mathbbm{1}_{c})$, where $\bm{w} \in \mathbb{R}^{M}$ and $\mathcal{W}(\cdot)$ consists of a multi-layer perceptron (MLP) and a normalization layer, which can construct diverse targeted vectors $\bm{w}$ for a given target class $c$. Thus $\mathcal{W}$ is capable of learning effective targeted latent vectors by randomly sampling different classes $c \in \mathcal{C}$ in training phase.

\textbf{Generator.} Given an input image $\bm{x}_{s}$, the encoder first calculates the feature map $\bm{F} \in \mathbb{R}^{N\times H \times W}$, where $N$, $H$ and $W$ refer to the number of channels, height and width of the feature map, respectively. The target latent vector $\bm{w}$, derived from the mapping network $\mathcal{W}$ by introducing a specific target class $c$, is expanded along height and width directions to obtain the label feature map $\bm{w}_s\in \mathbb{R}^{M\times H \times W}$. Then the above two feature maps are concatenated along the channels to obtain $\bm{F}' \in \mathbb{R}^{(N+M)\times H \times W}$. The obtained mixed feature map is then fed to the subsequent network.
Therefore, our generator $\mathcal{G}_{\theta}$ translates an input image $\bm{x}_{s}$ and latent target vector $\bm{w}$ into an output image $\mathcal{G}_{\theta}(\bm{x}_{s}, \bm{w})$, which enables $\mathcal{G}_{\theta}$ to synthesize adversarial images of a series of targets. For the output of feature map $\bm{f} \in \mathbb{R}^{d}$ in the decoder, we adopt a \textbf{smooth projection} $P(\cdot)$ to perform a change of variables over $\bm{f}$ rather than directly minimizing its $\ell_2$ norm as~\cite{han2019once} or clipping values outside the fixed norm~\cite{naseer2019cross}, which can be denoted as 
\begin{equation}
\label{eq:project}
   \bm{\delta} = P(\bm{f}) = \epsilon \cdot \mathrm{tanh}(\bm{f}),
\end{equation}
where $\epsilon$ is the strength of perturbation. Since $-1 \leq \mathrm{tanh}(\bm{f}) \leq 1$, $\bm{\delta}$ can automatically satisfy the $\ell_{\infty}$-ball bound with perturbation budget $\epsilon$. This transformation can be regarded as a better smoothing of gradient than directly clipping values outside the fixed norm, which is also instrumental for $\mathcal{G}_{\theta}$ to probe and learn the targeted semantic knowledge from $\mathcal{F}_{\phi}$.


\textbf{Training objectives.}
The training objectives seek to minimize the classification error on the perturbed image of the generator as 
\begin{equation}
    \theta^{*} \leftarrow \argmin_{\theta}{\mathbb{CE} \Big(F_{\phi}\big(\bm{x}_{s} + \mathcal{G}_{\theta}(\bm{x}_{s}, \mathcal{W}(\mathbbm{1}_{c}))\big), c\Big)},
\label{eq:objective}
\end{equation}
which adopts an end-to-end training paradigm with the goal of generating adversarial images to mislead the classifier the target label, and $\mathbb{CE}$ is the cross entropy loss. Previous studies attempt different classification losses in their works~\cite{zhang2020understanding,naseer2019cross}, and we found that cross-entropy loss works well in our settings. The detailed optimization procedure is summarized in Algorithm~\ref{algo1}.

\begin{algorithm}[t]
    \caption{Training Algorithm for the Conditional Generative Attack}\label{algo1}
\begin{algorithmic}[1]
\Require Training Data $\mathcal{D}_{s}$; a generative network $\mathcal{G}_{\theta}$; a classification network $\mathcal{F}_{\phi}$; a mapping network $\mathcal{W}$.
\Ensure Adversarial perturbations $\bm{\theta}$.
\For{iter in MaxIterations T}
\State Randomly sample $B$ images $\{\bm{x}_{s_{i}}\}_{i=1}^{B}$;
\State Randomly sample $B$ target classes $\{{c}_{i}\}_{i=1}^{B}$;
\State Forward pass ${c}_{i}$ into $\mathcal{W}$ to compute the targeted latent vectors $\bm{w}_{i}$;
\State Obtain the perturbed images by $\bm{x}_{s_{i}}^{*} = \epsilon \cdot \mathrm{tanh}(\mathcal{G}(\bm{x}_{s_{i}}, \bm{w}_{i})) + \bm{x}_{s_{i}}$;
\State Forward pass $\bm{x}_{s_{i}}^{*}$ to $\mathcal{F}_{\phi}$ and compute loss in Eq.~\eqref{eq:objective};
\State Backward pass and update the $\mathcal{G}_{\theta}$;

\EndFor
        
\end{algorithmic}
\end{algorithm}

\subsection{Hierarchical Partition for Classes} 

While handling plenty of classes, the effectiveness of a conditional generative model will decrease as illustrated in Fig.~\ref{fig:numbers}, because the representative capacity is limited with a single generator. Therefore, we propose to divide all classes into a feasible number of subsets to train models when the class number $K$ is large, e.g., 1,000 classes in ImageNet, with the aim of seeking the effectiveness of targeted black-box attack. To obtain a good partition, we introduce a representative target class space, which is nearly equivalent to the original class space $\mathcal{C}$. Specifically, we utilize the weights $\phi_{cls} \in \mathbb{R}^{D\times K}$ in the classifier layer for the
classification network $\mathcal{F}_{\phi}$. Therefore, $\phi_{cls}$ can be regarded as the alternative class space since the weight vector $\bm{d}_{c} \in \mathbb{R}^{D}$ from $\phi_{cls}$ can represent a class center of the feature embeddings of input images with same class $c$. 

Note that once those subsets with closer metric distance (e.g., larger cosine similarity) in the target class space $\phi_{cls}$ are regarded as conditional inputs of generative network, they obtain worse loss convergence and transferability than diverse them due to mutual influence among these input conditions, as illustrated in Fig.~\ref{fig:loss}. Thus we focus on selecting target classes that do not tend to overlap or be close to each other as accessible subsets. To capture more diverse examples in a given sampling space, we adopt K-determinantal point processes (DPP)~\cite{kulesza2012determinantal,kulesza2011kdpp} to achieve a hierarchical partition, which can take advantage of the diversity property among subsets by assigning subset probabilities proportional to determinants
of a kernel matrix.

First, we compute the RBF kernel matrix $L$ of $\phi_{cls}$ and eigendecomposition of $L$, and a random subset $V$ of the eigenvectors is chosen by regarding the eigenvalues as sampling probability. Second, we select a new class $c_{i}$ to add to the set and update $V$ in a manner that de-emphaseizes items similar to the one  selected. Each successive point is selected
and $V$ is updated by Gram-Schmidt orthogonalization, and the distribution shifts to
avoid points near those already chosen. By performing the above procedure, we can obtain a subset with $k$ size. Thus while handling the conditional classes with K, we can hierarchically adopt this algorithm to get the final $K/k$ subsets, which are regarded as conditional variables of generative models to craft adversarial examples. The details are presented in Appendix {\color{red} A}.

\section{Experiments}

In this section, we present extensive experiments to demonstrate the effectiveness of proposed method for targeted black-box attacks\footnote{Code at \url{https://github.com/ShawnXYang/C-GSP}.}. 

\subsection{Experimental Settings}
\label{sec:settings}

\textbf{Datasets.}
We consider the following datasets for training, including a widely used object detection dataset MS-COCO~\cite{lin2014microsoft} and ImageNet training set~\cite{deng2009imagenet}. We focus on standard and comprehensive testing settings, thus the inference is performed on ImageNet validation set (50k samples), a subset (5k) of ImageNet proposed by~\cite{li2019regional} and ImageNet-NeurIPS (1k) proposed by~\cite{nips2017}.

\textbf{Networks.}
We consider some naturally trained networks, i.e., Inception-v3 (Inv3)~\cite{szegedy2016rethinking}, Inception-v4 (Inv4)~\cite{szegedy2016inception}, Resnet-v2-152 (R152)~\cite{He2016} and Inception-Resnet-v2 (IR-v2)~\cite{szegedy2016inception}, which are widely used for evaluating transferability. Besides, we supplement DenseNet-201 (DN)~\cite{huang2017densely}, GoogleNet (GN)~\cite{szegedy2015going} and VGG-16 (VGG)~\cite{simonyan2014very} to fully evaluate the transferability.
Some adversarially trained networks~\cite{tramer2017ensemble} are also selected to evaluate the performance, i.e.,  ens3-adv-Inception-v3 ($\textrm{Inv3}_\textrm{ens3}$), ens4-adv-Inception-v3 ($\textrm{Inv3}_\textrm{ens4}$) and ens-adv-Inception-ResNet-v2 ($\textrm{IR-v2}_\textrm{ens}$). 



\textbf{Implementation details.} As for instance-specific attacks, we compare our method with several attacks, including MIM~\cite{Dong2017}, DIM~\cite{xie2019improving}, TI~\cite{dong2019evading}, SI~\cite{lin2019nesterov} and the state-of-the-art targeted attack named Logit~\cite{zhao2021success}. All instance-specific attacks adopt optimal hyperparameters provided in their original work.
Specifically, the attack iterations $M$ of MIM, DIM and Logit are set as $10, 20, 300$, respectively. And $\|W\|_{1}$ = 5 is used for TI~\cite{dong2019evading} as suggested by~\cite{gao2020patch,zhao2021success}. We choose the same ResNet autoencoder architecture in~\cite{johnson2016perceptual,naseer2019cross} as the basic generator networks, which consists of downsampling, residual and upsampling layers. We initialize the learning rate
as 2e-5 and set the mini-batch size as 32.
Smoothing mechanism is proposed to improve the transferability against adversarially trained models~\cite{dong2019evading}. Instead of adopting smoothing for generated perturbation while the training is completed as CD-AP~\cite{naseer2019cross}, we introduce adaptive Gaussian smoothing kernel to compute $\bm{\delta}$ from Eq.~\eqref{eq:project} in the training phase, named \textbf{adaptive Gaussian smoothing}, with the focus of making generated results obtain adaptive ability.  More implementation details and discussion with other networks (e.g., BigGAN~\cite{brock2018large}) are illustrated in Appendix {\color{red} B}. 

\begin{table*}[t]
    \begin{center}
    \scriptsize
    \setlength{\tabcolsep}{1pt}
    \caption{Transferability comparison for  multi-target attacks on ImageNet NeurIPS validation set (1k images) with the perturbation budget of $\ell_{\infty} \leq 16$. The results are averaged on 8 different target classes. Note that CD-AP$^{\dagger }$ indicates that training \textbf{8 models} can obtain results, while our method only train \textbf{one} conditional generative model. * indicates white-box attacks. }
    \label{tab:transfer}
    \begin{tabular}{p{4ex}<{\centering}|c|cc|ccccccc|ccc}
    \hline
        \multirow{2}{*}{} & \multirow{2}{*}{Method} & \multirow{2}{*}{\tabincell{c}{Time\\(ms)}} & \multirow{2}{*}{\tabincell{c}{Model\\ Number}} & \multicolumn{7}{c|}{\emph{Naturally Trained}} & \multicolumn{3}{c}{\emph{Adversarially Trained}} \\
         & & &&Inv3 & Inv4 & IR-v2 & R152 & DN & GN & VGG-16 & $\textrm{Inv3}_\textrm{ens3}$ & $\textrm{Inv3}_\textrm{ens4}$ & $\textrm{IR-v2}_\textrm{ens}$  \\
         \hline
         \hline
         \multirow{9}{*}{{\rotatebox[origin=c]{90}{Inv3}}}& MIM &$\sim$130&-& 99.9$^*$ & 0.8 & 1.0 & 0.4 & 0.2 & 0.2 &0.3 &$<$0.1 & 0.1 &$<0.1$ \\ & TI-MIM &$\sim$130&-& 99.9$^*$ & 0.9 & 1.1 & 0.4 & 0.4& 0.3 & 0.5 & 0.1 & 0.2 & 0.1\\ & SI-MIM &$\sim$130& - & 99.8$^*$ & 1.5 & 2.0 & 0.8 & 0.7 & 0.7 &0.5 & 0.3 & 0.3 & 0.1\\
         & DIM &$\sim$260&-&95.6$^*$ & 4.0 & 4.8 & 1.3 & 1.9 & 0.8 & 1.3 & 0.1 & 0.2 & 0.1\\ & TI-DIM&$\sim$260&-&96.0$^*$ & 4.4 & 5.1 & 1.4 & 2.4& 1.1 & 1.8 & 0.3 & 0.4 & 0.2\\ &
        SI-DIM &$\sim$260&-&98.4$^*$ & 5.6 & 5.9 & 2.8 & 3.0& 2.3 & 1.6& 0.9 & 0.9 & 0.3\\
        & Logit &$\sim$3900& - & 99.6$^*$ & 5.6 & 6.5 & 1.7 & 3.0 & 0.8 & 1.5 & 0.2 & 0.3 & 0.1 \\
        \cline{2-14}
        & CD-AP$^{\dagger }$ &$\sim${\color{blue} 15}& 8 &94.2$^*$ & 57.6 & 60.1 & 37.1 & 41.6 & 32.3 &41.7 & 1.5 & 2.2 & 1.2 \\ & CD-AP-gs$^{\dagger }$ &$\sim${\color{blue} 15}& 8 & 69.7$^*$ & 31.3 & 30.8 & 18.6 & 20.1 & 14.8 & 20.2 & 5.0 & 5.8 & 4.5 \\ &
        Ours &$\sim${\color{blue} 15}& {\color{blue} 1} & 93.4$^*$ & \textbf{66.9} & \textbf{66.6} & \textbf{41.6} & \textbf{46.4} & \textbf{40.0} & \textbf{45.0} & \textbf{39.7} & \textbf{37.2} & \textbf{32.2} \\
        \hline
        \hline
        \multirow{9}{*}{{\rotatebox[origin=c]{90}{R152}}}& MIM &$\sim$185&-& 0.5 & 0.4 & 0.6 & 99.7$^*$ & 0.3 & 0.3 &0.2 &0.1 & 0.1 &$<0.1$ \\ & TI-MIM &$\sim$185&-& 0.3 & 0.3 & 1.0 & 96.5$^*$ &  0.5& 0.3& 0.3  & 0.3 & 0.2 & 0.3 \\ & SI-MIM&$\sim$185&-&1.3 & 1.2 & 1.6 & 99.5$^*$ & 1.0 & 1.4 &0.7 & 0.3 & 0.4 & 0.2\\
         & DIM &$\sim$370&-&2.8 & 3.1 & 5.0 & 93.6$^*$ & 3.5 & 1.7 & 1.3 & 0.4 & 0.4 & 0.3\\ & TI-DIM&$\sim$370&-&4.3 & 4.1 & 5.8 & 92.9$^*$ & 4.3 & 2.1 & 1.4 & 0.8 & 0.7 & 0.4\\ &
        SI-DIM &$\sim$370&-& 7.2 & 8.4 & 10.4 & 97.4$^*$ & 7.6& 6.4 & 2.6 & 0.8 & 0.7 & 1.3\\
        & Logit &$\sim$5550 & - &  10.1 & 10.7 & 12.8 & 95.7$^*$ & 12.4 & 3.7 & 3.5 & 1.1 & 0.9 & 0.4 \\
        \cline{2-14}
        & CD-AP$^{\dagger }$ &$\sim${\color{blue} 10}& 8 &33.3 & 43.7 & 42.7 & 96.6$^*$ & 53.8 & 36.6 &34.1 & 15.7 & 15.2 & 12.0 \\ & CD-AP-gs$^{\dagger }$ &$\sim${\color{blue} 10}& 8 & 7.8 & 11.3 & 10.0 & 53.6$^*$ & 20.4 & 8.7 & 12.5 & 4.9 & 6.4 & 6.2 \\ &
        Ours &$\sim${\color{blue} 10}& {\color{blue} 1} & \textbf{37.7} & \textbf{47.6} & \textbf{45.1} & 93.2$^*$ & \textbf{64.2} & \textbf{41.7} & \textbf{45.9} & \textbf{31.6} & \textbf{32.0} & \textbf{29.9} \\
        \hline
         \hline
    \end{tabular}
    \end{center}
    
\end{table*}

\begin{figure}[t]
\begin{center}
\includegraphics[width=0.99\linewidth]{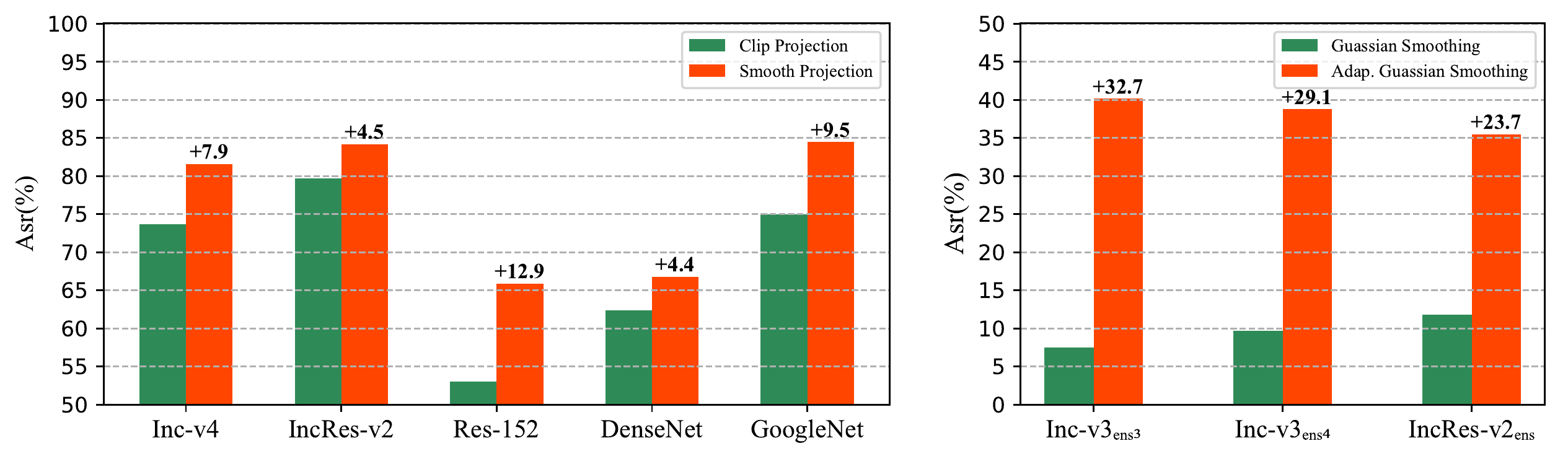}
\end{center}
\caption{Comparison of different projection functions and modes of Gaussian Smoothing. Results are reported with Inv3 network
on ImageNet NeurIPS validation set.}
\label{fig:ablation}
\end{figure}

\begin{table}[t]
    \begin{center}
    \caption{The untargeted fooling ratio (UT-FR) and targeted fooling ratio (T-FR) for adversarial attacks on ImageNet validation set (50k images) with the perturbation budget of $\ell_{\infty} \leq 10$. The attack is performed in same setting~\cite{zhang2020understanding} with the target class `sea lion' and the training dataset MS-COCO. * indicates white-box attacks.}
    \label{tab:transfer2}
    \footnotesize 
    \setlength{\tabcolsep}{5pt}
    \begin{tabular}{c|c|cc|cc|cc}
    \hline
       \multirow{2}{*} & \multirow{2}{*}{Method} & \multicolumn{2}{c|}{VGG-16} & \multicolumn{2}{c|}{VGG-19} & \multicolumn{2}{c}{R152} \\
       \cline{3-8}
       & & UT-FR & T-FR & UT-FR & T-FR & UT-FR & T-FR \\
       \cline{1-8}
         {VGG-16} &  \tabincell{c}{UAE~\cite{zhang2020understanding} \\ Ours} & \tabincell{c}{93.62$^*$\\\textbf{95.30}$^*$} &\tabincell{c}{82.90$^*$\\\textbf{83.54}$^*$} & \tabincell{c}{82.99\\\textbf{90.13}} & \tabincell{c}{13.69\\\textbf{38.59}} & \tabincell{c}{\textbf{36.03}\\35.15} & \tabincell{c}{0.01\\\textbf{0.14}}\\  
         \cline{1-8}
         {VGG-19} & \tabincell{c}{UAE~\cite{zhang2020understanding} \\ Ours} & \tabincell{c}{83.40\\\textbf{88.20}} & \tabincell{c}{44.53\\\textbf{48.96}} & \tabincell{c}{92.53$^*$\\\textbf{92.69}$^*$} & \tabincell{c}{\textbf{75.61}$^*$\\73.96$^*$} & \tabincell{c}{35.36\\\textbf{35.96}}  & \tabincell{c}{0.01\\\textbf{0.14}}\\
         \cline{1-8}
         {R152} & \tabincell{c}{UAE~\cite{zhang2020understanding} \\ Ours} &\tabincell{c}{55.05\\\textbf{83.90}} & \tabincell{c}{1.63\\\textbf{29.81}} & \tabincell{c}{55.12\\\textbf{83.24}} & \tabincell{c}{1.05\\\textbf{24.81}} & \tabincell{c}{82.58$^*$\\\textbf{91.14}$^*$} & \tabincell{c}{70.20$^*$\\\textbf{80.47}$^*$} \\ 
         \hline
    \end{tabular}
    \end{center}

\end{table}

\subsection{Transferability Evaluation}

We consider 8 different target classes from~\cite{zhang2020understanding} to form the multi-target black-box attack testing protocol with 8k times in 1k ImageNet NeurIPS set.

\textbf{Efficiency of multi-target black-box attack.} Among comparable methods, instance-specific methods, i.e.,  MIM, DIM, and Logit, require iterative mechanism with $M$ steps by computing gradients to obtain adversarial examples. Given the cost $t_{C}^{FP}$ and $t_{C}^{BP}$ of forward and backward passing the classifier, computing cost $T^{IS}$ of single data can be defined as $T^{IS} = t_{C}^{FP} * M + t_{C}^{BP} * M$ in Table~\ref{tab:transfer}. Instance-agnostic methods only require the inference cost from the trained generator as $T^{IA} = t_{G}^{FP}$, thus possessing the 
priority for those attack scenarios within limited time. However, instance-agnostic methods require to train 8 models to obtain all predictions from 8 different classes. Due to time-consuming training and more storage, we only reproduce an excellent generative method CD-AP~\cite{naseer2019cross} as a baseline, which already fully demonstrate the superior performance than other generative methods such as GAP~\cite{poursaeed2018generative} in their work. As a comparison, our conditional generative method only trains one model to inference the results and outperforms other methods w.r.t \emph{efficiency}. 

\textbf{Effectiveness of multi-target black-box attack.} Table~\ref{tab:transfer} shows the transferability comparison of different methods on both naturally and adversarially trained models. The success rate of instance-specific attacks are very unsatisfactory, possibly explained by the data-point overfitting that makes it hard to transfer another model. The instance-agnostic attack CD-AP obtains acceptable performance, yet inferior to proposed method w.r.t black-box transferability. 
The \textbf{primary reason} for such a trend lies in some distinctions as 1) direct clip projection in CD-AP and our smooth projection in Eq.~\eqref{eq:project} and 2) their Gaussian Smoothing and our adaptive Gaussian Smoothing, as described in  Sec.~\ref{sec:settings} and Appendix {\color{red} B}. Fig.~\ref{fig:ablation} empirically shows the comparison results of single-target black-box attacks based on the CD-AP framework. Thus proposed conditional generative method can be a reliable baseline w.r.t targeted black-box attacks, regarding both \emph{effectiveness} and \emph{efficiency}.

\textbf{Results of single-target black-box attack.} Recent related works, e.g., UAE~\cite{zhang2020understanding} and TTP~\cite{naseer2021generating} report excellent single-target black-box performances based on universal perturbations or functions. We obtain single-target degraded version of our model by specifying an input target label during the training process. The performance of black-box targeted attack between different methods is presented in Table~\ref{tab:transfer2}. Besides, we also make some analyses about TTP  and present compared results in Appendix {\color{red} C}.
Furthermore, some other instance-agnostic adversarial methods, e.g., UAP~\cite{moosavi2017universal}, GAP~\cite{poursaeed2018generative} and RHP~\cite{li2019regional}, have tendency towards the untargeted black-box problem. Despite this, we also follow the corresponding untargeted setting and compare these methods in Appendix {\color{red} C}. Our method is steadily improved under black-box targeted and untargeted black-box manner.


\begin{table}[t]
    \begin{center}
    \footnotesize 
    \setlength{\tabcolsep}{2pt}
    \caption{Transferability comparison with the perturbation budget of $\ell_{\infty} \leq 16$. White-box substitute model is Inv3 for all attacks, following the standard protocol~\cite{dong2019evading} with \textbf{1,000 stochastic target classes}.}
    \label{tab:transfer3}
    
    \begin{tabular}{p{11ex}<\raggedright | p{10ex}<\centering p{10ex}<\centering p{10ex}<\centering p{10ex}<\centering p{10ex}<\centering p{13ex}<\centering}
    \hline
    \multicolumn{7}{c}{\emph{Targeted Black-box  Attack in NeurIPS 2017 Competition ({\color{blue} 1,000 target classes})}} \\
    \hline
    \hline
        {Method} & Inv4 & IRv2 & R152 & DN & GN & VGG-16\\
        \hline
         MIM & 0.1 &  $<$0.1 & $<$0.1 & 0.3 & 0.1 &  $<$0.1\\ 
         TI-MIM & 0.3 & 0.3 & $<$0.1 & 0.4 & $<$0.1 & 0.1\\ 
         SI-MIM & 0.6 & 0.6 & 0.1 & 0.4 & 0.3 & 0.1\\
         DIM &  2.9 & 2.5 & 0.6 & 1.2 & 0.2 &  0.6\\
         TI-DIM & 2.9 & 2.5 & 0.5 & 1.7 & 0.3 & 1.0 \\ 
         SI-DIM & 4.3 &4.1 & 1.7 & 1.9 & 1.8 & 1.1\\
         Logit & 4.7 & 2.4 & 1.2 & 2.4& 0.4 & 0.8\\
         Ours &\textbf{35.9} & \textbf{37.4} & \textbf{25.0} & \textbf{26.8} & \textbf{22.9} & \textbf{26.6} \\  
         
         \hline
    \end{tabular}
    \end{center}
    
\end{table}

\begin{table}[t]
    \begin{center}
    \footnotesize 
    \setlength{\tabcolsep}{8pt}
    \caption{The success rate of black-box  \emph{impersonation} attacks on face verification with the perturbation budget of $\ell_{\infty} \leq 16$. ArcFace is chosen as white-box model.}
    \label{tab:transfer-face}
    
    \begin{tabular}{c|c|cccc}
    \hline
    \multicolumn{6}{c}{\emph{Black-box Impersonation Attack in Face Recognition}} \\
    \hline
    \hline
        {Protocol} & {Method}  & FaceNet & CosFace & SphereFace & MobileFace \\
         \cline{1-6}
         {\emph{I}}& \tabincell{c}{MIM \\  DIM \\ Ours} & \tabincell{c}{34.4\\38.8\\ \textbf{65.2}} & \tabincell{c}{16.6\\21.2\\ \textbf{56.2}} & \tabincell{c}{22.4\\27.4 \\ \textbf{52.2}} & \tabincell{c}{35.0\\44.3 \\ \textbf{83.5}} \\
         
         \cline{1-6}
         {\emph{II}}& \tabincell{c}{MIM \\  DIM \\ Ours} & \tabincell{c}{31.3\\36.1\\ \textbf{66.8}} &
         \tabincell{c}{13.6\\16.4\\ \textbf{49.1}} &
         \tabincell{c}{21.1\\24.4\\ \textbf{47.9}} &
         \tabincell{c}{22.3\\31.9\\ \textbf{67.8}} \\  

         \hline
    \end{tabular}
    \end{center}
    
\end{table}

\begin{figure*}[t]
\begin{center}
\includegraphics[width=0.99\linewidth]{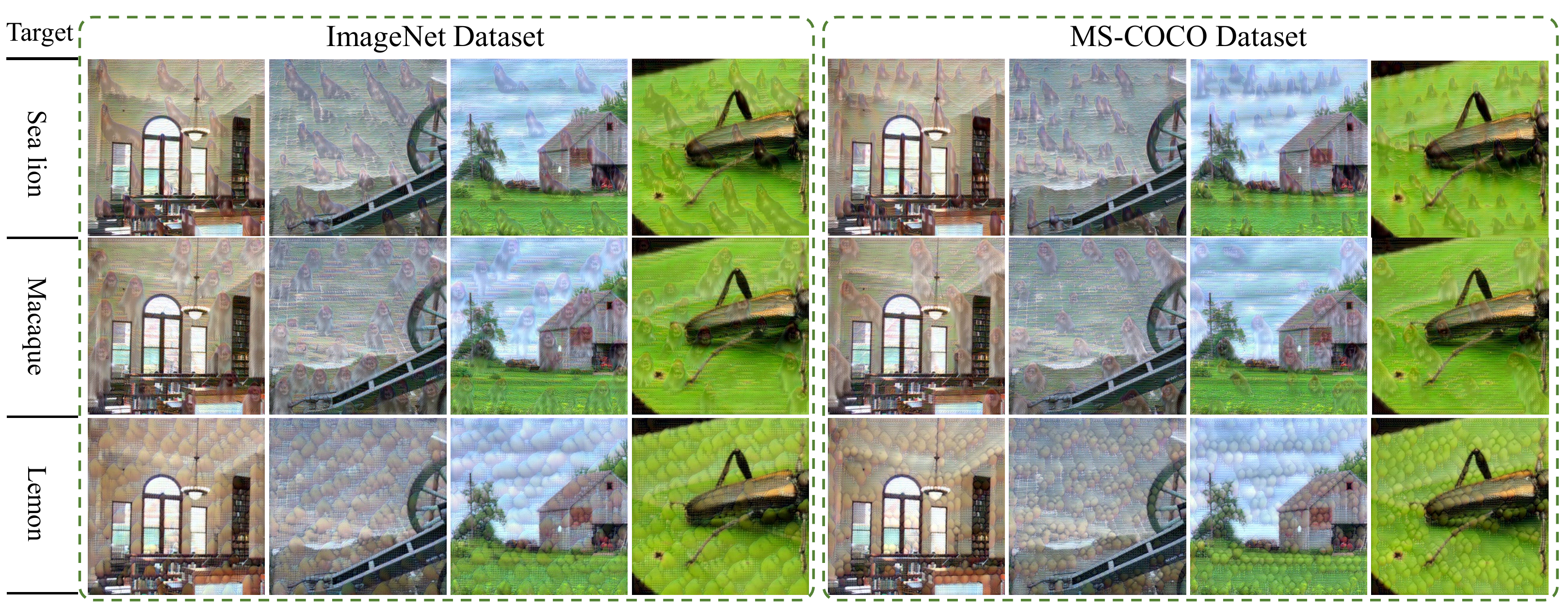}
\end{center}
\caption{Generative examples of adversarial images with perturbation budget of $\ell_{\infty}\leq 16$. We separately adopt the ImageNet and MS-COCO dataset as the training dataset to implement the generation of targeted perturbations. Our method can generate semantic pattern independent of training dataset.}
\label{fig:examples}

\end{figure*}

\begin{figure*}[t]
  \begin{minipage}[t]{.48\linewidth}
  \centering
  \includegraphics[width=0.99\textwidth]{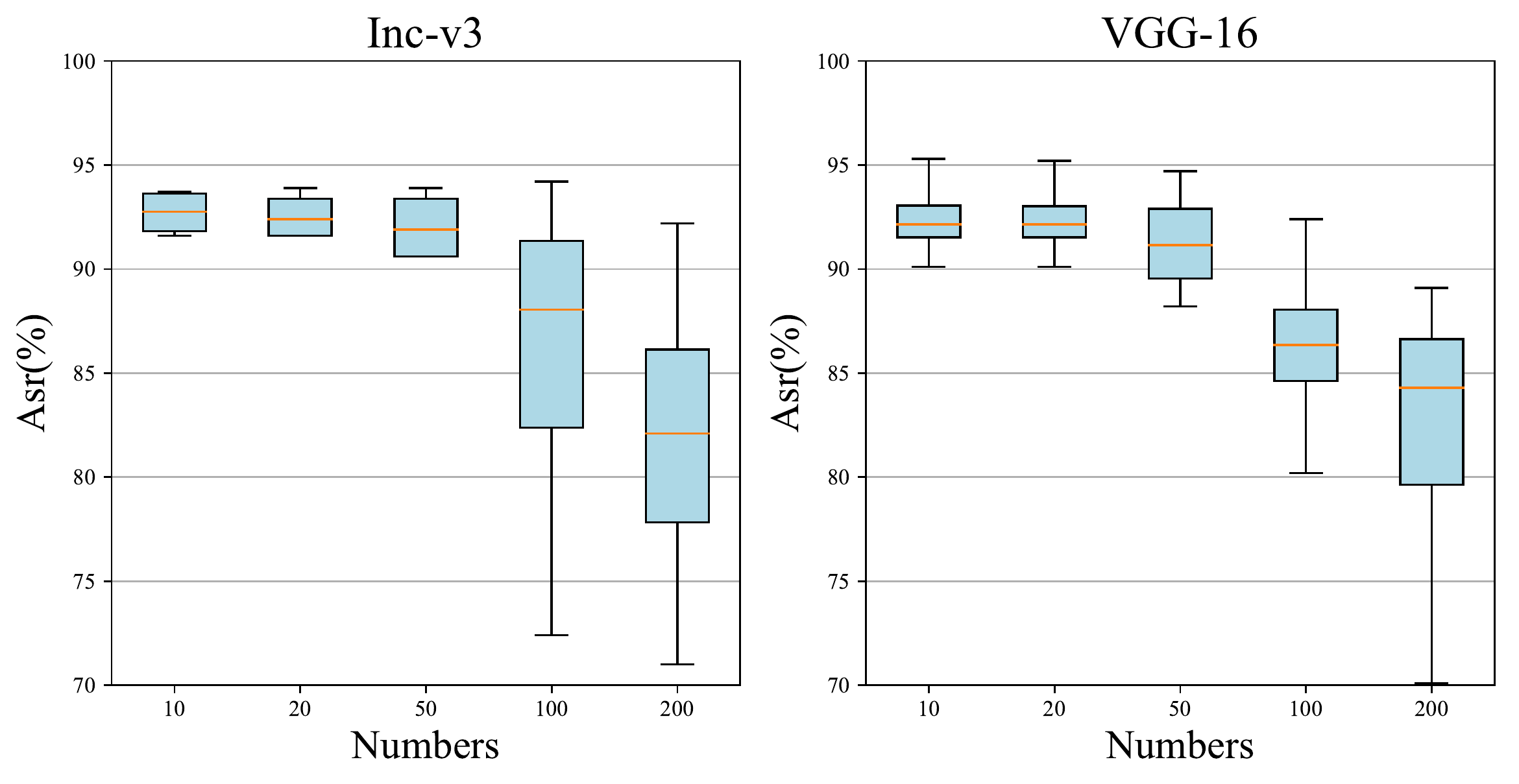}
  \captionof{figure}{Asr vs. \emph{numbers of conditional targets} curve against  Inv3 and VGG-16 models.}
\label{fig:numbers}
\end{minipage}
\begin{minipage}[t]{.48\linewidth}
  \centering
  \includegraphics[width=0.99\textwidth]{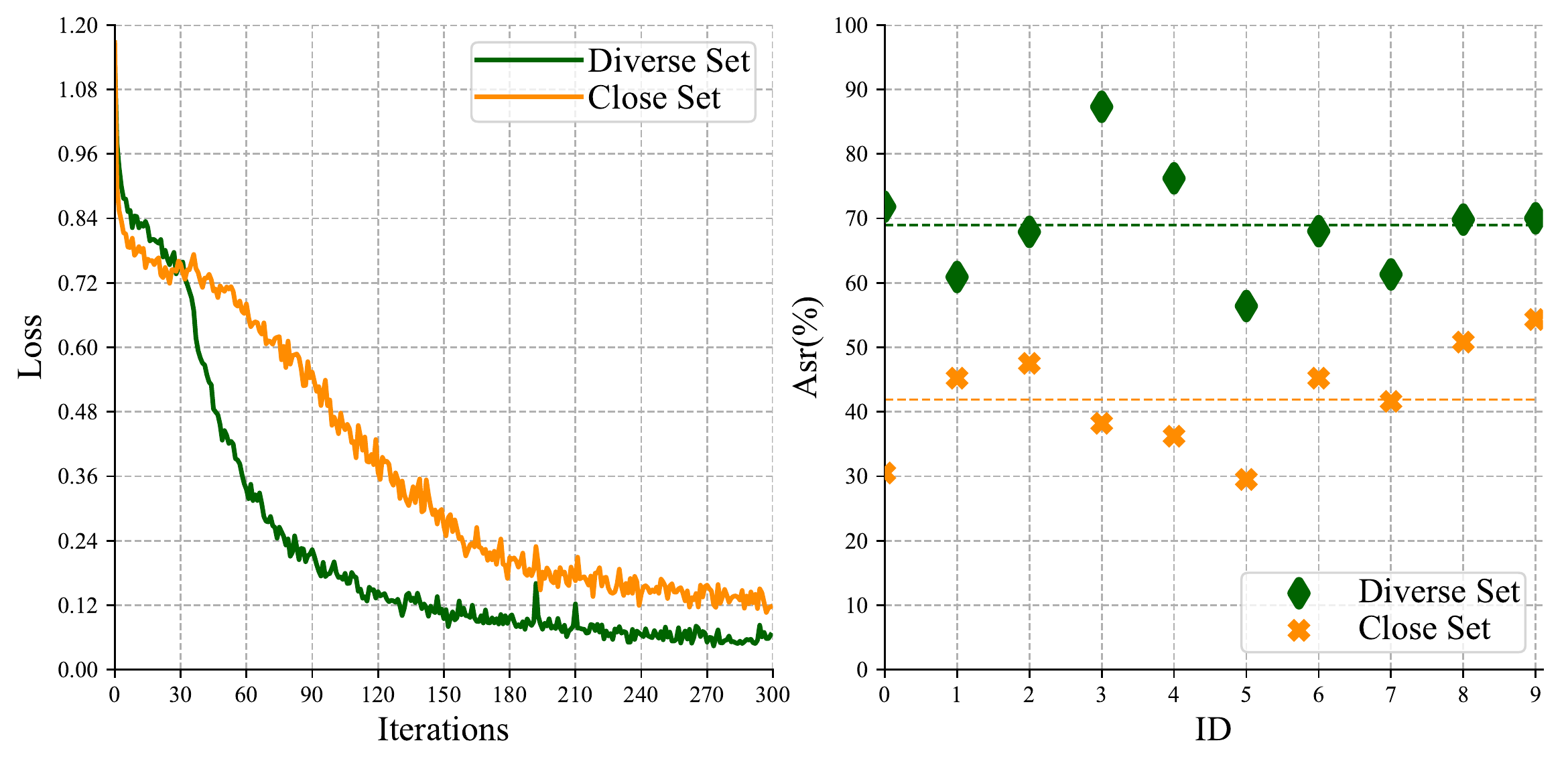}
  \captionof{figure}{Comparison of loss convergence and transferability between diverse and close conditional subset.}
\label{fig:loss}
\end{minipage}
  \end{figure*}

\subsection{Effectiveness on NeurIPS 2017 Competition}

To illustrate the effectiveness of our proposed attack methods in practical 1,000 classification, we here follow the official setting from NeurIPS 2017 adversarial competition~\cite{kurakin2018adversarial} for testing targeted black-box transferability. Considering limited resource, previous instance-agnostic attacks are not required as comparable methods due to training 1,000 models, thus we focus on various excellent instance-specific attacks for comparison, including official top attack methods in NeurIPS 2017 adversarial competition. Compared with other instance-agnostic attacks, our hierarchical partition mechanism can make conditional generative networks be capable of specifying any target class via a feasible number of models for the scalability. Specifically, we consider 20 models, with each specifying 50 diverse classes from k-DPP hierarchical partition in this setting, to implement targeted attack by only once inference for each target image. As shown in Table~\ref{tab:transfer3}, our method can obviously outperform all other methods. In addition, these trained generative models can directly be applied to craft adversarial examples, which is more convenient and efficient when required to handle large-scale
(e.g., millions of images) datasets than instance-specific attacks.

\subsection{Effectiveness on Realistic Face Recognition}

Adversarial perturbations added to original face images have ability to evade being recognized or impersonate another individual~\cite{sharif2016accessorize,yang2020delving}. In this section, we consider the transferability of impersonation attack to further illustrate the generalization of our method, which is also corresponding to targeted attack in the image classification task.

\textbf{Dataset and models.}
We conduct the experiments on Labeled Faces in the Wild (LFW)~\cite{huang2008labeled} and introduce two test protocols. For \emph{Protocol I} defined as single-target impersonation attack, we choose 1 target identity and 1k source face images belonging with different identities from LFW as the attackers, thus forming 1k pairs. For \emph{Protocol II} named multi-target impersonation attack, 5 target identities and 1k source face images are selected to form 1k attack pairs, meaning that we need to implement 5k attacks. We involve some excellent face recognition models for conducting black-box testing, including Sphereface~\cite{liu2017sphereface}, CosFace~\cite{wang2018cosface}, FaceNet~\cite{schroff2015facenet} and MobileFace~\cite{chen2018mobilefacenets}. These models lie in different model architectures and training objectives. In all experiments, we only use one model ArcFace~\cite{deng2019arcface} as substitute model to craft adversarial samples, and test attack performance against other unknown models.

\begin{figure*}[t]
  \begin{minipage}[t]{.53\linewidth}
  \centering
  \includegraphics[width=0.99\textwidth]{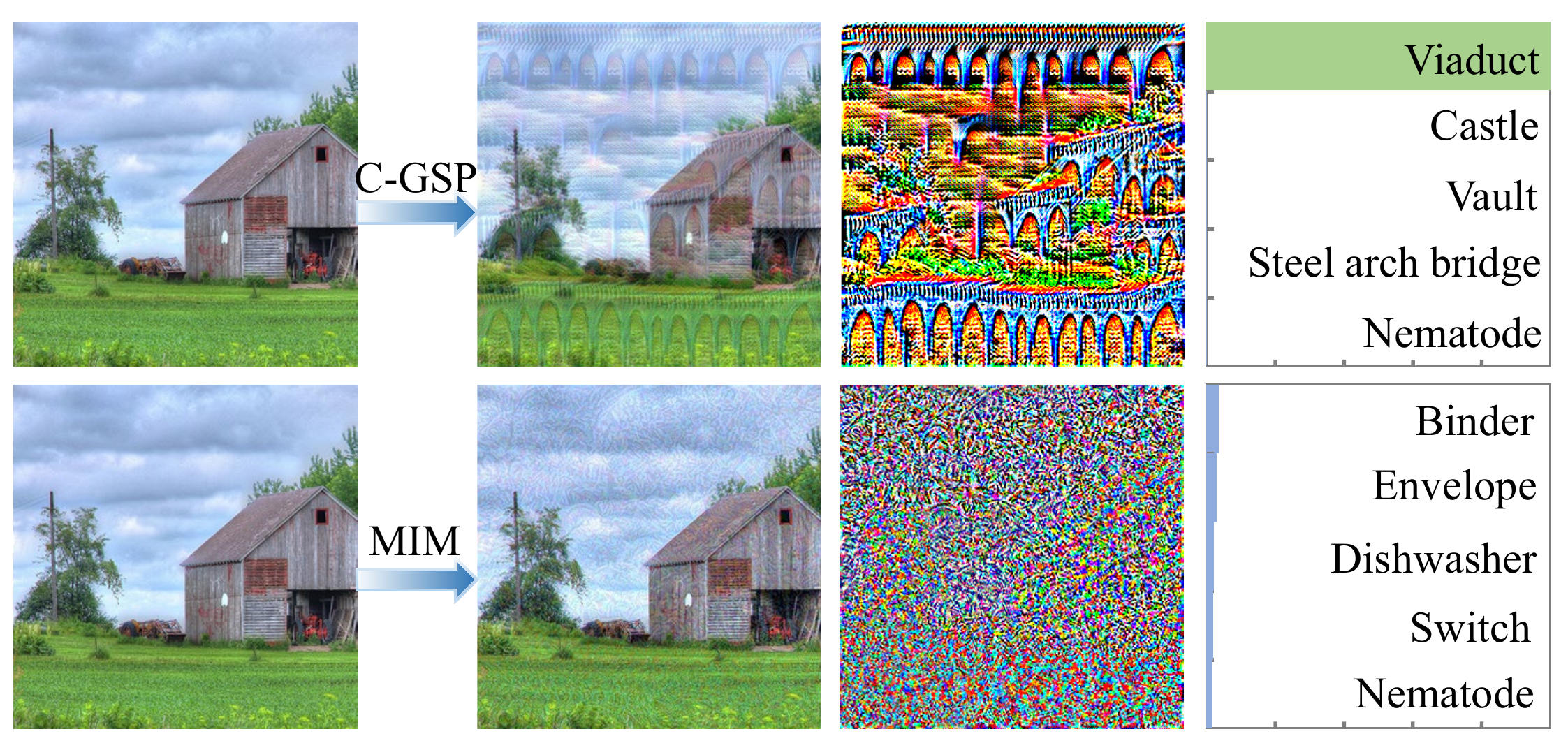}
  \captionof{figure}{We show the adversarial examples and extracted perturbation scaled in image-pixel space in the second column and the third column. The predictive confidence is presented by \textbf{directly feeding} extracted perturbation into the classifier in the last column.}
\label{fig:compare}
\end{minipage}
\begin{minipage}[t]{.47\linewidth}
  \centering
  \includegraphics[width=0.99\textwidth]{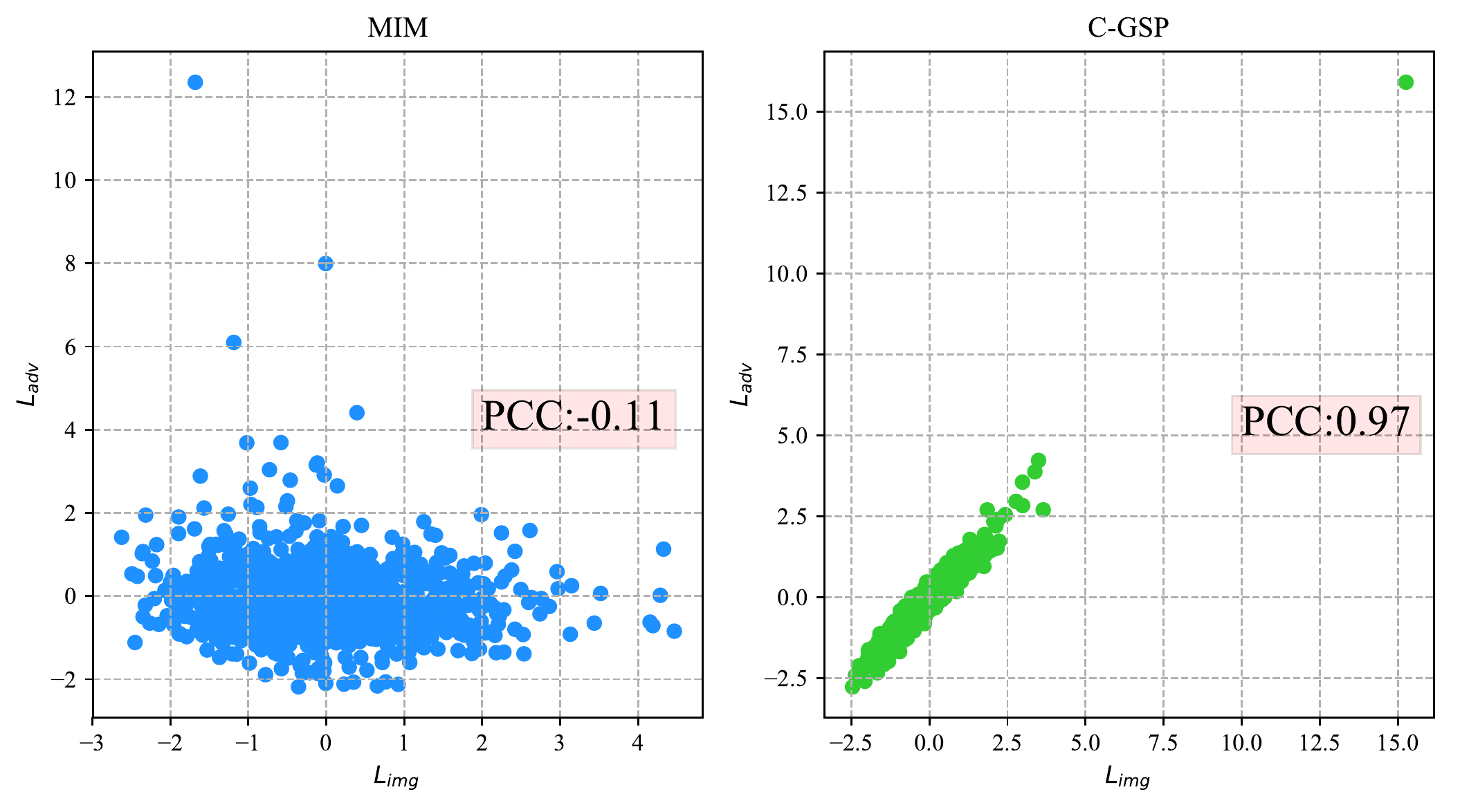}
  \captionof{figure}{Plots of logit vectors from the adversarial image $L_{img}$ and scaled crafted perturbation $L_{adv}$ of MIM and proposed generative method, with their respective PCC values.}
\label{fig:pcc}
\end{minipage}
  \end{figure*}

\textbf{Evaluation metrics.} We first compute the optimal threshold of every face recognition models from LFW dataset by following standard protocols. If the similarity of a pair of images exceeds the threshold, we regard them as same identity, otherwise different identities. 

\textbf{Black-box attack results.} 
We adjust the optimization object function to adapt face recognition for chosen attack methods (detailed in Appendix {\color{red} C}), and report the success rate of black-box \emph{impersonation} attacks in Table~\ref{tab:transfer-face}, which illustrates that our method can achieve nearly two times of the success rates than DIM in \emph{Protocol I} and \emph{Protocol II}. The results indicate that our method is superior to other methods not only in image classification.

\subsection{Comparison Study about Target Classes}

\label{sec:comparison}
We conduct an extensive study to investigate two key points about target classes. 

\textbf{Different numbers of target classes.} We conduct effectiveness for different numbers of target classes in Fig.~\ref{fig:numbers}. It can be seen that the results perform well within a feasible number of targets, whereas to a certain extent effectiveness tend to decay.
Therefore, the effectiveness of conditional generative networks is influenced by the number of conditional classes, due to the representative capacity of single generator. We aim to divide all classes into a feasible number of set while handling plenty of classes.

\textbf{Comparison of different multi-target conditions.}
We select closer conditional classes with larger cosine similarity in the target class space $\phi_{cls}$ and diverse conditional classes from k-DPP method. In Fig.~\ref{fig:loss}, closer conditional classes have worse loss convergence and transferability than diverse them due to mutual influence among conditions.

\subsection{More Analyses}

\label{lab:discuss}
Targeted adversarial samples from proposed generative method can produce semantic pattern inherent to the target class in Fig.~\ref{fig:examples}. Why does generative semantic pattern work? 

First, \emph{generative methods can produce strong targeted semantic pattern that is robust to the influence of data}, which is obtained by minimizing the loss of specific target class in the training phase. To corroborate our claim, we directly feed scaled crafted perturbations by instance-specific attack MIM and our generative method into the classifier. Indeed, we find that our generative perturbation is considered as target class with a high degree of confidence whereas the perturbation from MIM performs like the noise, as shown in Fig.~\ref{fig:compare}. Furthermore, we plot the logit relationship by computing PCC (Pearson correlation coefficient) values from scaled crafted perturbation and adversarial image in Fig.~\ref{fig:pcc}. The numerical performance is also consistent with our mentioned claim.

Second, \emph{the generated adversarial semantic pattern achieves well-generalizing performance among the different models}. We feed 1k images from ImageNet test set into the generator trained by Inv3 model to obtain 1k semantic patterns, which are scaled to image pixel space and then fed into different classifiers. We compute the mean confidence of $\bm{0.46}$ for DN, $\bm{0.44}$ for Inv4, and $\bm{0.35}$ for R152, whereas the perturbation from MIM is lower than $0.01$. The results show that our scaled semantic pattern can directly achieve well-generalizing performance among models, possibly explained by utilizing similar feature knowledge from the same class on different classifiers trained on same training data distribution. Thus similar pattern can be instrumental for transferability among models.

\section{Discussion and Conclusion}

Transferability of targeted black-box attack is simultaneously affected by data and model. Therefore, instance-specific methods easily overfit the data point and white-box model, resulting in weak transferability. As a comparison, the proposed generative method with powerful learning capacity reduces the dependency for data point by adopting the unlabeled training data, thus enabling the model to learn semantic pattern and improve the transferability of targeted black-box attack. Extensive experiments demonstrate that proposed generative method can significantly improve the success rates of targeted black-box attacks against various models, meanwhile achieving faster speedup beyond an order of magnitude than gradient-based methods. Therefore, this method can be regarded as a new baseline method in terms of targeted black-box attacks, which provides a novel framework to explore the vulnerabilities of DNNs.

\medskip
{
\noindent{\textbf{Acknowledgement}}. This work was supported by the National Key Research and Development Program of China (Nos. 2020AAA0104304, 2017YFA0700904), NSFC Projects (Nos. 62061136001, 61621136008, 62076147, U19B2034, U19A2081, U1811461), the major key project of PCL (No. PCL2021A12), Tsinghua-Alibaba Joint Research Program,  Tsinghua-OPPO Joint Research Center, and the High Performance Computing Center, Tsinghua University.}

%
%

\bibliographystyle{splncs04}
\bibliography{egbib}

\begin{thebibliography}{10}
\providecommand{\url}[1]{\texttt{#1}}
\providecommand{\urlprefix}{URL }
\providecommand{\doi}[1]{https://doi.org/#1}

\bibitem{berthelot2017began}
Berthelot, D., Schumm, T., Metz, L.: Began: Boundary equilibrium generative
  adversarial networks. arXiv preprint arXiv:1703.10717  (2017)

\bibitem{comics}
BircanoAYlu, C.:
  \url{https://www.kaggle.com/cenkbircanoglu/comic-books-classification.
  Kaggle}. Kaggle, 2017

\bibitem{brock2018large}
Brock, A., Donahue, J., Simonyan, K.: Large scale gan training for high
  fidelity natural image synthesis. arXiv preprint arXiv:1809.11096  (2018)

\bibitem{chen2018mobilefacenets}
Chen, S., Liu, Y., Gao, X., Han, Z.: Mobilefacenets: Efficient cnns for
  accurate real-time face verification on mobile devices. In: Chinese
  Conference on Biometric Recognition. pp. 428--438. Springer (2018)

\bibitem{demontis2019adversarial}
Demontis, A., Melis, M., Pintor, M., Jagielski, M., Biggio, B., Oprea, A.,
  Nita-Rotaru, C., Roli, F.: Why do adversarial attacks transfer? explaining
  transferability of evasion and poisoning attacks. In: 28th $\{$USENIX$\}$
  Security Symposium ($\{$USENIX$\}$ Security 19). pp. 321--338 (2019)

\bibitem{deng2009imagenet}
Deng, J., Dong, W., Socher, R., Li, L.J., Li, K., Fei-Fei, L.: Imagenet: A
  large-scale hierarchical image database. In: 2009 IEEE conference on computer
  vision and pattern recognition. pp. 248--255. Ieee (2009)

\bibitem{deng2019arcface}
Deng, J., Guo, J., Xue, N., Zafeiriou, S.: Arcface: Additive angular margin
  loss for deep face recognition. In: Proceedings of the IEEE Conference on
  Computer Vision and Pattern Recognition. pp. 4690--4699 (2019)

\bibitem{dong2019benchmarking}
Dong, Y., Fu, Q.A., Yang, X., Pang, T., Su, H., Xiao, Z., Zhu, J.: Benchmarking
  adversarial robustness. In: The IEEE Conference on Computer Vision and
  Pattern Recognition (CVPR) (2020)

\bibitem{Dong2017}
Dong, Y., Liao, F., Pang, T., Su, H., Zhu, J., Hu, X., Li, J.: Boosting
  adversarial attacks with momentum. In: Proceedings of the IEEE Conference on
  Computer Vision and Pattern Recognition (CVPR) (2018)

\bibitem{dong2019evading}
Dong, Y., Pang, T., Su, H., Zhu, J.: Evading defenses to transferable
  adversarial examples by translation-invariant attacks. In: Proceedings of the
  IEEE Conference on Computer Vision and Pattern Recognition (CVPR) (2019)

\bibitem{eykholt2018robust}
Eykholt, K., Evtimov, I., Fernandes, E., Li, B., Rahmati, A., Xiao, C.,
  Prakash, A., Kohno, T., Song, D.: Robust physical-world attacks on deep
  learning visual classification. In: IEEE Conference on Computer Vision and
  Pattern Recognition. pp. 1625--1634 (2018)

\bibitem{gao2020patch}
Gao, L., Zhang, Q., Song, J., Liu, X., Shen, H.T.: Patch-wise attack for
  fooling deep neural network. In: European Conference on Computer Vision. pp.
  307--322. Springer (2020)

\bibitem{Goodfellow-et-al2016}
Goodfellow, I., Bengio, Y., Courville, A.: Deep Learning. MIT Press (2016),
  \url{http://www.deeplearningbook.org}

\bibitem{goodfellow2014explaining}
Goodfellow, I.J., Shlens, J., Szegedy, C.: Explaining and harnessing
  adversarial examples. In: International Conference on Learning
  Representations (ICLR) (2015)

\bibitem{han2019once}
Han, J., Dong, X., Zhang, R., Chen, D., Zhang, W., Yu, N., Luo, P., Wang, X.:
  Once a man: Towards multi-target attack via learning multi-target adversarial
  network once. In: Proceedings of the IEEE International Conference on
  Computer Vision. pp. 5158--5167 (2019)

\bibitem{He2016}
He, K., Zhang, X., Ren, S., Sun, J.: Identity mappings in deep residual
  networks. In: European Conference on Computer Vision (ECCV). pp. 630--645.
  Springer (2016)

\bibitem{hendrycks2021unsolved}
Hendrycks, D., Carlini, N., Schulman, J., Steinhardt, J.: Unsolved problems in
  ml safety. arXiv preprint arXiv:2109.13916  (2021)

\bibitem{huang2017densely}
Huang, G., Liu, Z., Van Der~Maaten, L., Weinberger, K.Q.: Densely connected
  convolutional networks. In: Proceedings of the IEEE conference on computer
  vision and pattern recognition. pp. 4700--4708 (2017)

\bibitem{huang2008labeled}
Huang, G.B., Mattar, M., Berg, T., Learned-Miller, E.: Labeled faces in the
  wild: A database forstudying face recognition in unconstrained environments.
  In: Technical report (2007)

\bibitem{huang2019enhancing}
Huang, Q., Katsman, I., He, H., Gu, Z., Belongie, S., Lim, S.N.: Enhancing
  adversarial example transferability with an intermediate level attack. In:
  Proceedings of the IEEE/CVF International Conference on Computer Vision. pp.
  4733--4742 (2019)

\bibitem{inkawhich2020perturbing}
Inkawhich, N., Liang, K., Wang, B., Inkawhich, M., Carin, L., Chen, Y.:
  Perturbing across the feature hierarchy to improve standard and strict
  blackbox attack transferability. Advances in Neural Information Processing
  Systems  \textbf{33},  20791--20801 (2020)

\bibitem{inkawhich2020transferable}
Inkawhich, N., Liang, K.J., Carin, L., Chen, Y.: Transferable perturbations of
  deep feature distributions. arXiv preprint arXiv:2004.12519  (2020)

\bibitem{johnson2016perceptual}
Johnson, J., Alahi, A., Fei-Fei, L.: Perceptual losses for real-time style
  transfer and super-resolution. In: European conference on computer vision.
  pp. 694--711. Springer (2016)

\bibitem{kulesza2011kdpp}
Kulesza, A., Taskar, B.: k-dpps: Fixed-size determinantal point processes. In:
  ICML (2011)

\bibitem{kulesza2012determinantal}
Kulesza, A., Taskar, B.: Determinantal point processes for machine learning.
  arXiv preprint arXiv:1207.6083  (2012)

\bibitem{Kurakin2016}
Kurakin, A., Goodfellow, I., Bengio, S.: Adversarial examples in the physical
  world. In: International Conference on Learning Representations (ICLR)
  Workshops (2017)

\bibitem{kurakin2018adversarial}
Kurakin, A., Goodfellow, I., Bengio, S., Dong, Y., Liao, F., Liang, M., Pang,
  T., Zhu, J., Hu, X., Xie, C., et~al.: Adversarial attacks and defences
  competition. In: The NIPS'17 Competition: Building Intelligent Systems, pp.
  195--231. Springer (2018)

\bibitem{li2020towards}
Li, M., Deng, C., Li, T., Yan, J., Gao, X., Huang, H.: Towards transferable
  targeted attack. In: Proceedings of the IEEE/CVF Conference on Computer
  Vision and Pattern Recognition. pp. 641--649 (2020)

\bibitem{li2019regional}
Li, Y., Bai, S., Xie, C., Liao, Z., Shen, X., Yuille, A.L.: Regional
  homogeneity: Towards learning transferable universal adversarial
  perturbations against defenses. arXiv preprint arXiv:1904.00979  (2019)

\bibitem{lin2019nesterov}
Lin, J., Song, C., He, K., Wang, L., Hopcroft, J.E.: Nesterov accelerated
  gradient and scale invariance for adversarial attacks. In: International
  Conference on Learning Representations (2019)

\bibitem{lin2014microsoft}
Lin, T.Y., Maire, M., Belongie, S., Hays, J., Perona, P., Ramanan, D.,
  Doll{\'a}r, P., Zitnick, C.L.: Microsoft coco: Common objects in context. In:
  European conference on computer vision. pp. 740--755. Springer (2014)

\bibitem{liu2017sphereface}
Liu, W., Wen, Y., Yu, Z., Li, M., Raj, B., Song, L.: Sphereface: Deep
  hypersphere embedding for face recognition. In: Proceedings of the IEEE
  conference on computer vision and pattern recognition. pp. 212--220 (2017)

\bibitem{liu2016delving}
Liu, Y., Chen, X., Liu, C., Song, D.: Delving into transferable adversarial
  examples and black-box attacks. In: ICLR (2017)

\bibitem{moosavi2017universal}
Moosavi-Dezfooli, S.M., Fawzi, A., Fawzi, O., Frossard, P.: Universal
  adversarial perturbations. In: Proceedings of the IEEE conference on computer
  vision and pattern recognition. pp. 1765--1773 (2017)

\bibitem{Moosavidezfooli2016}
Moosavi-Dezfooli, S.M., Fawzi, A., Frossard, P.: Deepfool: a simple and
  accurate method to fool deep neural networks. In: Proceedings of the IEEE
  Conference on Computer Vision and Pattern Recognition (CVPR) (2016)

\bibitem{naseer2019cross}
Naseer, M.M., Khan, S.H., Khan, M.H., Khan, F.S., Porikli, F.: Cross-domain
  transferability of adversarial perturbations. In: Advances in Neural
  Information Processing Systems. pp. 12905--12915 (2019)

\bibitem{naseer2021generating}
Naseer, M., Khan, S., Hayat, M., Khan, F.S., Porikli, F.: On generating
  transferable targeted perturbations. In: Proceedings of the IEEE/CVF
  International Conference on Computer Vision. pp. 7708--7717 (2021)

\bibitem{nips2017}
NeurIPS:
  \url{https://www.kaggle.com/c/nips-2017-defense-against-\\adversarial-attack/data}.
  Kaggle, 2017

\bibitem{poursaeed2018generative}
Poursaeed, O., Katsman, I., Gao, B., Belongie, S.: Generative adversarial
  perturbations. In: Proceedings of the IEEE Conference on Computer Vision and
  Pattern Recognition. pp. 4422--4431 (2018)

\bibitem{reddy2018ask}
Reddy~Mopuri, K., Krishna~Uppala, P., Venkatesh~Babu, R.: Ask, acquire, and
  attack: Data-free uap generation using class impressions. In: Proceedings of
  the European Conference on Computer Vision (ECCV). pp. 19--34 (2018)

\bibitem{schroff2015facenet}
Schroff, F., Kalenichenko, D., Philbin, J.: Facenet: A unified embedding for
  face recognition and clustering. In: Proceedings of the IEEE conference on
  computer vision and pattern recognition. pp. 815--823 (2015)

\bibitem{sharif2016accessorize}
Sharif, M., Bhagavatula, S., Bauer, L., Reiter, M.K.: Accessorize to a crime:
  Real and stealthy attacks on state-of-the-art face recognition. In:
  Proceedings of the 2016 ACM SIGSAC Conference on Computer and Communications
  Security. pp. 1528--1540 (2016)

\bibitem{simonyan2014very}
Simonyan, K., Zisserman, A.: Very deep convolutional networks for large-scale
  image recognition. arXiv preprint arXiv:1409.1556  (2014)

\bibitem{song2018constructing}
Song, Y., Shu, R., Kushman, N., Ermon, S.: Constructing unrestricted
  adversarial examples with generative models. In: Advances in Neural
  Information Processing Systems (NeurIPS) (2018)

\bibitem{szegedy2016inception}
Szegedy, C., Ioffe, S., Vanhoucke, V., Alemi, A.: Inception-v4,
  inception-resnet and the impact of residual connections on learning. In: AAAI
  (2017)

\bibitem{szegedy2015going}
Szegedy, C., Liu, W., Jia, Y., Sermanet, P., Reed, S., Anguelov, D., Erhan, D.,
  Vanhoucke, V., Rabinovich, A.: Going deeper with convolutions. In:
  Proceedings of the IEEE conference on computer vision and pattern
  recognition. pp.~1--9 (2015)

\bibitem{szegedy2016rethinking}
Szegedy, C., Vanhoucke, V., Ioffe, S., Shlens, J., Wojna, Z.: Rethinking the
  inception architecture for computer vision. In: Proceedings of the IEEE
  Conference on Computer Vision and Pattern Recognition (CVPR) (2016)

\bibitem{szegedy2013intriguing}
Szegedy, C., Zaremba, W., Sutskever, I., Bruna, J., Erhan, D., Goodfellow, I.,
  Fergus, R.: Intriguing properties of neural networks. In: International
  Conference on Learning Representations (ICLR) (2014)

\bibitem{tramer2017ensemble}
Tram{\`e}r, F., Kurakin, A., Papernot, N., Boneh, D., McDaniel, P.: Ensemble
  adversarial training: Attacks and defenses. In: International Conference on
  Learning Representations (ICLR) (2018)

\bibitem{wang2018cosface}
Wang, H., Wang, Y., Zhou, Z., Ji, X., Gong, D., Zhou, J., Li, Z., Liu, W.:
  Cosface: Large margin cosine loss for deep face recognition. In: Proceedings
  of the IEEE Conference on Computer Vision and Pattern Recognition. pp.
  5265--5274 (2018)

\bibitem{wu2020skip}
Wu, D., Wang, Y., Xia, S.T., Bailey, J., Ma, X.: Skip connections matter: On
  the transferability of adversarial examples generated with resnets. arXiv
  preprint arXiv:2002.05990  (2020)

\bibitem{xie2019improving}
Xie, C., Zhang, Z., Zhou, Y., Bai, S., Wang, J., Ren, Z., Yuille, A.L.:
  Improving transferability of adversarial examples with input diversity. In:
  Proceedings of the IEEE Conference on Computer Vision and Pattern Recognition
  (CVPR) (2019)

\bibitem{xu2019understanding}
Xu, K., Li, C., Zhu, J., Zhang, B.: Understanding and stabilizing gans'
  training dynamics with control theory. arXiv preprint arXiv:1909.13188
  (2019)

\bibitem{yang2022controllable}
Yang, X., Dong, Y., Pang, T., Xiao, Z., Su, H., Zhu, J.: Controllable
  evaluation and generation of physical adversarial patch on face recognition.
  arXiv e-prints pp. arXiv--2203 (2022)

\bibitem{yang2020towards}
Yang, X., Dong, Y., Pang, T., Zhu, J., Su, H.: Towards privacy protection by
  generating adversarial identity masks. arXiv preprint arXiv:2003.06814
  (2020)

\bibitem{yang2020design}
Yang, X., Wei, F., Zhang, H., Zhu, J.: Design and interpretation of universal
  adversarial patches in face detection. In: Computer Vision--ECCV 2020: 16th
  European Conference, Glasgow, UK, August 23--28, 2020, Proceedings, Part XVII
  16. pp. 174--191. Springer (2020)

\bibitem{yang2020delving}
Yang, X., Yang, D., Dong, Y., Yu, W., Su, H., Zhu, J.: Delving into the
  adversarial robustness on face recognition. arXiv preprint arXiv:2007.04118
  (2020)

\bibitem{yi2014learning}
Yi, D., Lei, Z., Liao, S., Li, S.Z.: Learning face representation from scratch.
  arXiv preprint arXiv:1411.7923  (2014)

\bibitem{zhang2020understanding}
Zhang, C., Benz, P., Imtiaz, T., Kweon, I.S.: Understanding adversarial
  examples from the mutual influence of images and perturbations. In:
  Proceedings of the IEEE/CVF Conference on Computer Vision and Pattern
  Recognition. pp. 14521--14530 (2020)

\bibitem{zhao2021success}
Zhao, Z., Liu, Z., Larson, M.: On success and simplicity: A second look at
  transferable targeted attacks. Advances in Neural Information Processing
  Systems  \textbf{34} (2021)

\end{thebibliography}
\clearpage
\appendix
\section{Sampling Algorithm}
We summarize the overall sampling procedure based on k-DPP~\cite{kulesza2011kdpp}  in Algorithm~\ref{algo2}. 

\begin{itemize}
    \item Compute the RBF kernel matrix $L$ of $\phi_{cls}$ and eigendecomposition of $L$.
    \item A random subset $V$ of the eigenvectors is chosen by regarding the eigenvalues as sampling probability.
    \item Select a new class $c_{i}$ to add to the set and update $V$ in a manner that de-emphaseizes items similar to the one  selected. 
    \item Update $V$ by Gram-Schmidt orthogonalization, and the distribution shifts to
avoid points near those already chosen.
\end{itemize}

By performing the Algorithm~\ref{algo2}, we can obtain a subset with $k$ size. Thus while handling the conditional classes with K, we can hierarchically adopt this algorithm to get the final $K/k$ subsets, which are regarded as conditional variables of generative models to craft adversarial examples.

\begin{algorithm}[t]
    \caption{Sampling Algorithm by kDPP}\label{algo2}
\begin{algorithmic}[1]
\Require{Weight Vector $\bm{\theta}_{cls}$; Subset size $k$.}
\Ensure{A subset $C$.}
\State Compute RBF kernel matrix $L$ of $\bm{\theta}_{cls}$;
\State Compute eigenvector/value $\{v_{n}, \lambda_{n}\}_{n=1}^N$ pairs of $L$;
\State // \emph{Phase I:}
\State    $J\leftarrow\phi$, $e_{k}\left(\lambda_{1}, \ldots, \lambda_{N}\right)=\sum_{|J|=k} \prod_{n \in J} \lambda_{n}$;
\For{n = N, ..., 1}
\If{$u\sim U[0,1] < \lambda_{n} \frac{e^{n-1}_{k-1}}{e^{n}_{k}}$ and $k>0$}
\State $J\leftarrow J \cup \{n\}$; $k\leftarrow k-1$;
\EndIf
\EndFor

\State // \emph{Phase II:}
\State $V \leftarrow\left\{v_{n}\right\}_{n \in J}, Y\leftarrow \phi$;
\While{$|V| > 0$}
\State Select $c_i$ from $\mathcal{C}$ with $\operatorname{P}\left(c_{i}\right)=\frac{1}{|V|} \sum_{v \in V}\left(v^{\top} e_{i}\right)^{2}$;
\State $C \leftarrow C \cup\left\{c_{i}\right\}$;
$V \leftarrow V_{\perp},$ an orthonormal basis for the subspace of $V$ orthogonal to $e_{i}$;
\EndWhile
\end{algorithmic}
\end{algorithm}

\begin{figure*}[t]
\begin{center}
\includegraphics[width=1.0\linewidth]{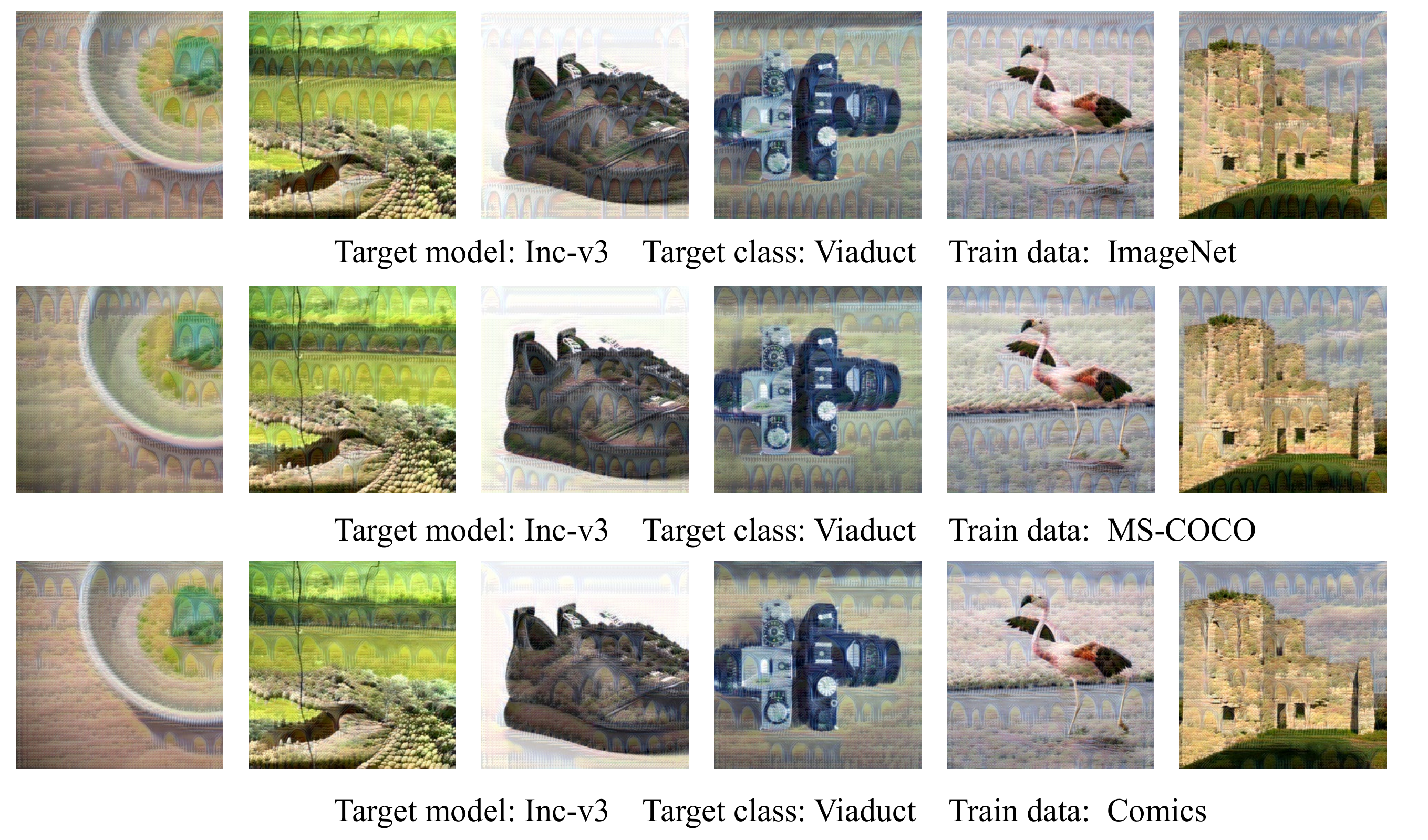}
\end{center}
\caption{Some examples of adversarial images with perturbation budget of $\ell_{\infty}\leq 16$. We separately adopt the ImageNet, MS-COCO and Comics dataset as the training dataset to implement the generation of targeted perturbations.}
\label{fig:a-dataset}
\end{figure*}

\section{Some Implementation Details}
\textbf{The study of smoothing mechanism.} Smoothing mechanism has been proved to improve the transferability against adversarially trained models. CD-AP~\cite{naseer2019cross} uses direct clip projection to have a fixed norm $\epsilon$, and adopts smoothing for generated perturbation while the generator $\mathcal{G}$ is trained, i.e., 
\begin{equation}
\label{eq:project-gs}
\begin{gathered}
   \textbf{Train: }\bm{x}_{s_{i}}^{*} = \mathrm{Clip}_{\epsilon}(\mathcal{G}(\bm{x}_{s_{i}}),\\
   \textbf{Test: }\bm{x}_{s_{i}}^{*} = \mathrm{W} * \mathrm{Clip}_{\epsilon}(\mathcal{G}(\bm{x}_{s_{i}}),
  \end{gathered}
\end{equation}
where $\mathrm{W}$ indicates Gaussian smoothing of kernel size of 3, $*$ indicates the convolution operation, and Clip$_{\epsilon}$ means clipping values outside the fixed norm $\epsilon$. As a comparison, we introduce adaptive Gaussian smoothing kernel to compute adversarial images $\bm{x}_{s_{i}}^{*}$ from in the training phase, named \textbf{adaptive Gaussian smoothing} as

\begin{equation}
\label{eq:project2}
   \textbf{Train \& Test: } \bm{x}_{s_{i}}^{*} = \epsilon \cdot \mathrm{W} * \mathrm{tanh}(\mathcal{G}(\bm{x}_{s_{i}}) + \bm{x}_{s_{i}},
\end{equation}
which can make generated results obtain adaptive ability in the training phase. 
We perform training in ImageNet dataset to report all results including comparable baselines.

\textbf{Network architecture of generator.} We adopt the same autoencoder architecture in~\cite{naseer2019cross} as the basic generator networks. Besides, we also explore BigGAN~\cite{brock2018large} as conditional generator network. An very weak testing performance is obtained even in the \emph{white-box} attack scenario, possibly explained by the weak diversity of latent variable with the Gaussian distribution from BigGAN in the training phase, whereas autoencoder can take full advantage of large-scale training dataset, e.g., ImageNet. Furthermore, we also train the autoencoder with Gaussian noise as the training dataset and obtain similar inferior performance in the white-box attack scenario, indicating that a large-scale training dataset is very significant for generating transferable targeted adversarial examples.


\section{Additional Experimental Results}
\textbf{Results on different datasets.} We craft adversarial examples on different datasets, including ImageNet training set, MS-COCO and Comics dataset~\cite{comics}, which consist of 1.2M, 82k and 50K images, respectively. MS-COCO  dataset can be applied to large-scale object detection and segmentation, and those images from Comics dataset are regarded as other domains different from normal ones in ImageNet. Despite this diverse training types, we still find the common property of crafted adversarial examples by our method. Specifically, we craft some examples of adversarial images with perturbation budget of $\ell_{\infty}\leq 16$, and separately adopt the ImageNet, MS-COCO and Comics dataset as the training dataset to implement the generation of targeted perturbations. As illustrated in Fig.~\ref{fig:a-dataset}, we produce semantic pattern independent of any training dataset.

\begin{table}[t]
    \begin{center}
    \small 
    \setlength{\tabcolsep}{10pt}
    \caption{Comparison results of targeted black-box attacks on different datasets. Inv3 is the substitute model.}
    \label{tab:data}
    \begin{tabular}{c|ccc}
    \hline
        Dataset & DN & VGG-16 & GN \\
         \hline
        ImageNet & 79.9 & 81.9 & 73.2 \\
     MS-COCO & 70.3 & 71.3 &64.1 \\ 
     Comics & 60.4 & 63.0 & 61.3 \\
         \hline
    \end{tabular}
    \end{center}
    
\end{table}

We also report the success rate of targeted black-box attack, as shown in Table~\ref{tab:data}. We experimentally find that semantic pattern derived from ImageNet dataset achieves better performance of black-box performance,  possibly explained by instructional effectiveness from more diverse data in ImageNet dataset.

\textbf{Results of untargeted black-box attack.} We evaluate our method and other generative methods including UAP~\cite{moosavi2017universal}, GAP~\cite{poursaeed2018generative} and RHP~\cite{li2019regional}. Untargeted transferability from naturally trained models to adversarially trained models occurs due to differences in model sources, data types and other factors, thus enabling challenging comparison.
As illustrated in Table~\ref{tab:transfer-un}, we report the untargeted attacks increase in error rate of adversarial and clean images to evaluate different methods.  Our method is steadily improved in different black-box models under untargeted black-box manner.

\textbf{Results of different $\epsilon$.} We also presented the results with {the reduced perturbation budget} of $\ell_{\infty}\leq 10$ in Table~\ref{tab:epsilon} for verifying the consistent effectiveness. Furthermore, we chose the smaller perturbation budget of $\ell_{\infty}\leq 8$ in experiments to make the adversarial examples more imperceptible. In this setting, the proposed generative method still outperforms the SOTA iterative attack method named Logit~\cite{zhao2021success} with a large margin. 

\textbf{Compared results with TTP~\cite{naseer2021generating}.} TTP proposes a generative approach for highly transferable targeted perturbations by introducing mutual distribution matching. For demonstrating the performance, we conduct multi-target black-box experiments by adopting 8 mutually exclusive targeted sets. 1) \emph{Efficiency}: TTP needs to train 8 models while performing an 8-class targeted attack, whereas our conditional generative method only trains one model to inference the results. 2) \emph{Effectiveness}: TTP obtained comparable black-box attack success rates with ours as shown in Table~\ref{tab:ttp}. Overall, the proposed conditional generative method can be a better baseline in targeted black-box attacks regarding both effectiveness and efficiency.

\begin{table}[t]
    \begin{center}
    \footnotesize 
    \setlength{\tabcolsep}{5pt}
    \caption{Transferability results for untargeted attacks increase in error rate after attack on subset of ImageNet (5k images) with the perturbation budget of $\ell_{\infty} \leq 16/32$.}
    \label{tab:transfer-un}
    \begin{tabular}{c|c|cc|cc|cc}
    \hline
       \multirow{2}{*} & \multirow{2}{*}{Method} & \multicolumn{2}{c|}{$\textrm{inv3}_\textrm{ens3}$} & \multicolumn{2}{c|}{$\textrm{inv3}_\textrm{ens4}$ } & \multicolumn{2}{c}{$\textrm{IR-v2}_\textrm{ens}$} \\
       \cline{3-8}
       & & $\epsilon = 16$ & $\epsilon = 32$ & $\epsilon = 16$ & $\epsilon = 32$ & $\epsilon = 16$ & $\epsilon = 32$ \\
       \cline{1-8}
         {inv3} &  \tabincell{c}{UAP~\cite{moosavi2017universal} \\ GAP~\cite{poursaeed2018generative} \\ RHP~\cite{li2019regional}} & \tabincell{c}{1.00 \\ 5.48 \\ 32.5} &\tabincell{c}{7.82 \\ 33.3 \\ 60.8} & \tabincell{c}{1.80 \\ 4.14 \\ 31.6} & \tabincell{c}{5.60 \\ 29.4 \\ 58.7} & \tabincell{c}{1.88 \\ 3.76 \\ 24.6} & \tabincell{c}{5.60 \\ 22.5 \\ 57.0}\\  
         \cline{1-8}
         {inv4} &  \tabincell{c}{UAP~\cite{moosavi2017universal} \\ RHP~\cite{li2019regional}} & \tabincell{c}{2.08 \\ 27.5} &\tabincell{c}{7.68 \\ 60.3} & \tabincell{c}{1.94 \\ 26.7} & \tabincell{c}{6.92 \\ 62.5} & \tabincell{c}{2.34 \\ 21.2} & \tabincell{c}{6.78 \\ 58.5}\\  
         \cline{1-8}
         {IR-v2} &  \tabincell{c}{UAP~\cite{moosavi2017universal} \\ RHP~\cite{li2019regional}} & \tabincell{c}{1.88 \\ 29.7} &\tabincell{c}{8.28 \\ 62.3} & \tabincell{c}{1.74 \\ 29.8} & \tabincell{c}{7.22 \\ 63.3} & \tabincell{c}{1.96 \\ 26.8} & \tabincell{c}{8.18 \\ 62.8}\\  
         \cline{1-8}
        \multicolumn{2}{c|}{CD-AP~\cite{naseer2019cross}} & 28.34 & 71.3 & 29.9 & 66.72 & 19.84 & 60.88 \\
         \multicolumn{2}{c|}{CD-AP-gs~\cite{naseer2019cross}} & 41.06 & 71.96 & 42.68 & 71.58 & 37.4 & 72.86 \\
         \multicolumn{2}{c|}{Ours} & \textbf{46.20} & \textbf{72.58} & \textbf{42.98} & \textbf{72.34} & \textbf{37.9} & \textbf{73.26} \\
        
         \hline
    \end{tabular}
    \end{center}

\end{table}

\begin{table}[t]
    \begin{center}
    \small 
    \setlength{\tabcolsep}{5pt}
    
    \caption{Comparison results of targeted black-box attacks on different $\epsilon$.}
    \label{tab:epsilon}
    \begin{tabular}{c|c|ccc|ccc}
    \hline
       \multirow{2}{*}{Source} & \multirow{2}{*}{Method} & \multicolumn{3}{c|}{VGG-16} & \multicolumn{3}{c}{R152} \\
       \cline{3-8}
       & & eps=16& eps=12& eps=8 & eps=16 & eps=12 & eps=8 \\
       \cline{1-8}
       \multirow{2}{*}{inv3} & Logit~\cite{zhao2021success} &  4.4& 3.4 & 2.1 & 1.2 & 1.1 & 0.8\\
       & Ours & 61.9 &  53.7 &  36.1 & 49.6 & 31.0  & 16.3 \\
       \hline
    \end{tabular}
    \end{center}

\end{table}

\begin{table}[t]
    \begin{center}
    \small 
    \setlength{\tabcolsep}{10pt}
    \caption{Comparison results of targeted black-box attacks with TTP.}
    \label{tab:ttp}
    \begin{tabular}{c|ccccccc}
    \hline
        Source & Method & Inv4 & IR-v2 & R152 & DN & GN & VGG-16\\
         \hline
        \multirow{2}{*}{inv3} & TTP & 65.4 & 55.3 & 39.4 & 44.0 & 35.9& 36.1\\
    & Ours & 66.9 & 66.6 & 41.6 & 46.4 & 40.0 & 45.0\\
         \hline
    \end{tabular}
    \end{center}
    
\end{table}

\section{Impersonation Attack of Face Recognition} We list attack methods of face recognition as follows.
Given an input $\bm{x}$ and an image $\bm{x}^r$ belonging with another identity, an attack method can generate an adversarial example $\bm{x}^{adv}$ with perturbation budget $\epsilon$ under the $\ell_p$ norm ($\|\bm{x}^{adv} - \bm{x}\|_p \leq \epsilon$). Therefore, impersonation attack aims to perform this objective of 

\begin{equation}\label{eq:delta}
    \mathcal{C} (\bm{x}^{adv}, \bm{x}^{r}) = \mathbb{I} (\mathcal{D}_f (\bm{x}^{adv}, \bm{x}^{r}) < \delta ),
\end{equation}
where $\mathbb{I}$ is the indicator function, $\delta$ is a threshold, and $\mathcal{D}_f(\bm{x}^{adv}, \bm{x}^{r}) = \|f(\bm{x}^{adv}) - f(\bm{x}^{r})\|_2^2.$

\textbf{Basic Iterative Method (BIM)}~\cite{Kurakin2016} extends FGSM by iteratively taking multiple small gradient updates as
\begin{equation}
\label{eq:bim}
    \bm{x}_{t+1}^{adv} = \mathrm{clip}_{\bm{x},\epsilon}  \big(\bm{x}_t^{adv} - \alpha\cdot\mathrm{sign}(\nabla_{\bm{x}}\mathcal{D}_f(\bm{x}_t^{adv},\bm{x}^{r}))\big),
\end{equation}
where $\mathrm{clip}_{\bm{x},\epsilon}$ projects the adversarial example to satisfy the $\ell_{\infty}$ constrain and $\alpha$ is the step size.

\textbf{Momentum Iterative Method (MIM)}~\cite{Dong2017} introduces a momentum term into BIM for improving the transferability of adversarial examples as 
\begin{equation}
\begin{gathered}
     \bm{g}_{t+1} = \mu \cdot \bm{g}_t + \frac{\nabla_{\bm{x}}\mathcal{D}_f(\bm{x}_t^{adv},\bm{x}^{r})}{\|\nabla_{\bm{x}}\mathcal{D}_f(\bm{x}_t^{adv},\bm{x}^{r})\|_1}; \\  \bm{x}^{adv}_{t+1}=\mathrm{clip}_{\bm{x},\epsilon}(\bm{x}^{adv}_t - \alpha\cdot\mathrm{sign}(\bm{g}_{t+1})).
\end{gathered}
\end{equation}

The training objectives of our generative method seek to minimize the classification error on the perturbed image of the generator as 
\begin{equation}
     \min_{\theta}\mathbb{E}_{(\bm{x}\sim \mathcal{X}, c\sim \mathcal{C})}[{\mathcal{D}_f \big(\bm{x} + \mathcal{G}_{\theta}(\bm{x}, c), \bm{x}^{r}_{c}\big)}],
\label{eq:a-objective}
\end{equation}
where $\bm{x}^{r}_{c}$ refers to $\bm{x}^{r}$ with the corresponding identity $c$. In the training phase, we randomly select $1,000$ identities from  CASIA-WebFace~\cite{yi2014learning} as training dataset to craft adversarial examples.  Therefore, our method can be applied not only in image classification.

\section{More Examples}

\begin{figure*}[t]
\begin{center}
\includegraphics[width=1.0\linewidth]{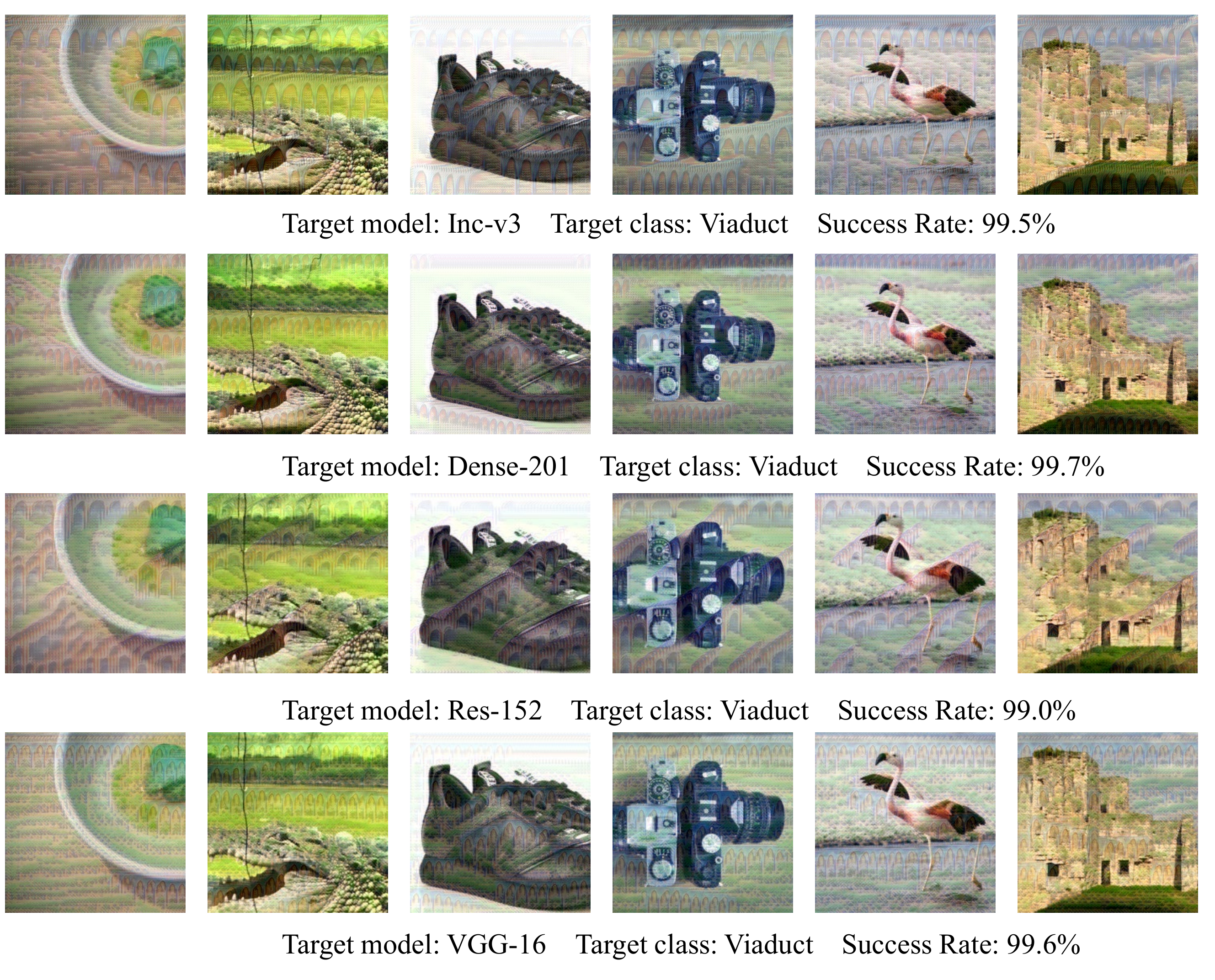}
\end{center}
\caption{Some examples of adversarial images with perturbation budget of $\ell_{\infty}\leq 16$. We separately adopt the ImageNet, MS-COCO and Comics dataset as the training dataset to implement the generation of targeted perturbations.}
\label{fig:a-model}
\end{figure*}

We also show more semantic patterns from different target models, as illustrated in Fig.~\ref{fig:a-model}.

\end{document}